\pdfoutput=1 
\documentclass{article}

\usepackage[final, nonatbib]{neurips_data_2022}

\usepackage[utf8]{inputenc} 
\usepackage[T1]{fontenc}    
\usepackage{hyperref}       
\usepackage{url}            
\usepackage{booktabs}       
\usepackage{amsfonts}       
\usepackage{nicefrac}       
\usepackage{microtype}      
\usepackage{xcolor}         

\usepackage{graphicx}
\usepackage{amsmath}
\usepackage{amssymb}
\usepackage{soul,color}
\usepackage{multirow}
\usepackage[flushleft]{threeparttable}
\usepackage{tabularx}
\usepackage{colortbl}
\usepackage[]{algorithm2e}
\usepackage{bbding}
\usepackage{enumitem}
\usepackage{arydshln}
\usepackage{caption}
\usepackage{subcaption}
\usepackage{wrapfig}
\usepackage{minitoc}
\usepackage{wrapfig}

\title{BLOX: Macro Neural Architecture Search Benchmark and Algorithms}

\author{%
  Thomas Chau\textsuperscript{1*}, Łukasz Dudziak\textsuperscript{1*}, Hongkai Wen\textsuperscript{1,3} \\
  \textbf{Nicholas D. Lane\textsuperscript{1,2}, Mohamed S. Abdelfattah\textsuperscript{4}} \\
  \\
  \textsuperscript{1} Samsung AI Center, Cambridge, UK\\
  \textsuperscript{2} University of Cambridge, UK \hspace*{10pt}
  \textsuperscript{3} University of Warwick, UK\\
  \textsuperscript{4} Cornell University, USA\\
  \footnotesize{\textsuperscript{*} \textit{Indicates equal contributions}} \\
  \\
  \texttt{\{thomas.chau, l.dudziak, hongkai.wen, nic.lane\}@samsung.com} \\
  \texttt{mohamed@cornell.edu}
}

\newcommand{\figvs}[5]{\begin{figure}[!t]
\centering
\includegraphics[width=#1\columnwidth,keepaspectratio,#3]{images_db1/#2}
\caption{#4}
\label{#5}
\end{figure}}

\newcommand{\figvsh}[5]{\begin{figure}[!h]
\centering
\includegraphics[width=#1\columnwidth,keepaspectratio,#3]{images_db1/#2}
\caption{#4}
\label{#5}
\end{figure}}

\newcommand{\figvsw}[5]{\begin{figure*}[!h]
\centering
\includegraphics[width=#1\columnwidth,keepaspectratio,#3]{images_db1/#2}
\caption{#4}
\label{#5}
\end{figure*}}

\newcommand{\blox}{Blox}
\newcommand{\comment}[1]{}

\newcounter{question}
\newcommand{\question}[1]{\refstepcounter{question}\par\noindent\textbf{Q\thequestion: #1}}

\makeatletter
\def\adl@drawiv#1#2#3{%
        \hskip.5\tabcolsep
        \xleaders#3{#2.5\@tempdimb #1{1}#2.5\@tempdimb}%
                #2\z@ plus1fil minus1fil\relax
        \hskip.5\tabcolsep}
\newcommand{\cdashlinelr}[1]{%
  \noalign{\vskip\aboverulesep
          }
  \cdashline{#1}
  \noalign{
          \vskip\belowrulesep}}
\makeatother

\begin{document}
\doparttoc 
\faketableofcontents 
\maketitle

\begin{abstract}
Neural architecture search (NAS) has been successfully used to design numerous high-performance neural networks.
However, NAS is typically compute-intensive, so most existing approaches restrict the search to decide the operations and topological structure of a single block only, then the same block is stacked repeatedly to form an end-to-end model.
Although such an approach reduces the size of search space, recent studies show that a macro search space, which allows blocks in a model to be different, can lead to better performance.
To provide a systematic study of the performance of NAS algorithms on a macro search space, we release \blox{} – a benchmark that consists of 91k unique models trained on the CIFAR-100 dataset.
The dataset also includes runtime measurements of all the models on a diverse set of hardware platforms.
We perform extensive experiments to compare existing algorithms that are well studied on cell-based search spaces, with the emerging blockwise approaches that aim to make NAS scalable to much larger macro search spaces.
The \blox{} benchmark and code are available at \url{https://github.com/SamsungLabs/blox}.

\end{abstract}

\vspace{-0.5cm}

\section{Introduction}\label{sec:intro}
Deep neural network (DNN) performance is closely related to its architecture topology and hyper-parameters as demonstrated through the progression of image classification CNNs in recent years: AlexNet~\cite{alexnet}, Inception~\cite{googlenet}, MobileNets~\cite{mobilenetv3} and EfficientNets~\cite{efficientnetv1,efficientnetv2}.
Increasingly, automated methods are used to design DNN architectures to avoid intuition-based manual design.
The field of neural architecture search (NAS) continues to offer a large number of methods including sample-based NAS~\cite{rlnas,evonas}, differentiable NAS~\cite{darts}, training-free NAS~\cite{zerocost} and blockwise NAS~\cite{dna2020,donna2021,hant}.
Within this realm of NAS for DNN design there are two important design problems which are still mostly manual.
First, how do we design the NAS \textit{search space}, which defines the set of DNN architectures from which a NAS algorithm can select.
Second, how do we select or design a suitable search method for a given NAS search space.
In this work, we attempt to address both problems through a focused analysis of \textit{macro} NAS algorithms within a new NAS search space called \blox{}.

\noindent \textbf{NAS search spaces.}
A well-defined search space is crucial for NAS. 
However, the literature has mostly focused on cell-based designs in which the NAS algorithm only searches for operations and connections of a cell that is repeatedly stacked within a predefined skeleton~\cite{darts, nasbench1, nasbench2, nao2018, npenas2020, bonas2020, bananas2020}.
These approaches prohibit layer diversity which can help to achieve both high accuracy and low latency~\cite{mnasnet2019}.
An alternative to cell-based NAS, known as macro NAS, enables the individual search for each block in a DNN as shown in Figure~\ref{fig:macro_vs_micro}.
In other words, macro NAS allows different stages of a model to have different structures.
Though promising, macro NAS is exorbitantly expensive because the search space size grows exponentially with the number of blocks.
We present the first large-scale benchmark and study of a macro search space to shed some light on how to perform NAS in this challenging setting.

\figvs{0.9}{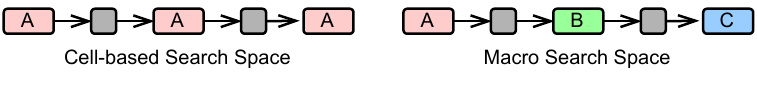}{trim=0 0 0 0}{A macro search space allows each block to have a different architecture whereas a cell-based search space repeats the same cell/block throughout the DNN.}{fig:macro_vs_micro}

\noindent \textbf{NAS benchmarks.}
To facilitate a fair comparison of NAS algorithms, standardized benchmarks have been created such as NAS-Bench-101/201/1shot1/NLP/ASR~\cite{nasbench1,nasbench2,nasbench1shot1,nasbenchnlp,nasbenchasr}. 
While these benchmarks span multiple application domains, they all use cell-based search spaces thus limiting the analysis of NAS algorithms to this setting only.
More recently, NAS-Bench-Macro~\cite{nbmacro} proposed a macro search space with 8 stages; however, each stage only has three block options making the overall search space quite small ($3^8=6,561$ DNNs) and not diverse.
To address this, we have developed \blox{} -- a much larger macro NAS benchmark that focuses on block diversity, with 45 unique block options and three stages ($45^3=91,125$ DNNs).
This enables the empirical analysis of NAS algorithms on macro search spaces and will thus inform the design of efficient macro search algorithms.
Table~\ref{tab:nasbenchmarks} summarizes Blox and other recent NAS benchmarks.

\noindent \textbf{Macro NAS algorithms.}
Any search algorithm can operate on a macro search space; however, very few will be efficient because of the large search space size.
To cope with the complexity of macro search, a new class of \textit{blockwise} search algorithms are being developed that perform local search within each stage before using that local information to construct an end-to-end model.
Blockwise search algorithms is a family of NAS algorithms designed to work well for macro NAS problems.
This divide-and-conquer approach has the potential to speed up macro NAS at the expense of using inexact heuristics to predict the performance of each block.
DNA~\cite{dna2020}, DONNA~\cite{donna2021} and LANA~\cite{hant} are three recent and notable works in this area, showing state-of-the-art accuracy-latency tradeoffs on very large macro search spaces.
In this work, we aim to analyze the different components of these blockwise NAS algorithms to understand, compare and build upon the existing approaches.

We enumerate our contributions below:

\begin{enumerate}[noitemsep]
    \item \textbf{Macro search space and benchmark for NAS.} We release \blox{}, a NAS benchmark for CNNs on a macro search space, trained on the CIFAR-100 dataset~\cite{cifar}, with latency measurements from multiple hardware devices.
    \item \textbf{Analysis of blockwise NAS.} We systematically evaluate the performance of different NAS algorithms on \blox{}, with a particular focus on emerging blockwise search algorithms, for which we include a detailed analysis of the efficacy of (a) block signatures, (b) accuracy predictors, and (c) training methodologies.
    
\end{enumerate}

\section{\blox{}: Macro Search Space}\label{sec:blox}

\figvs{0.8}{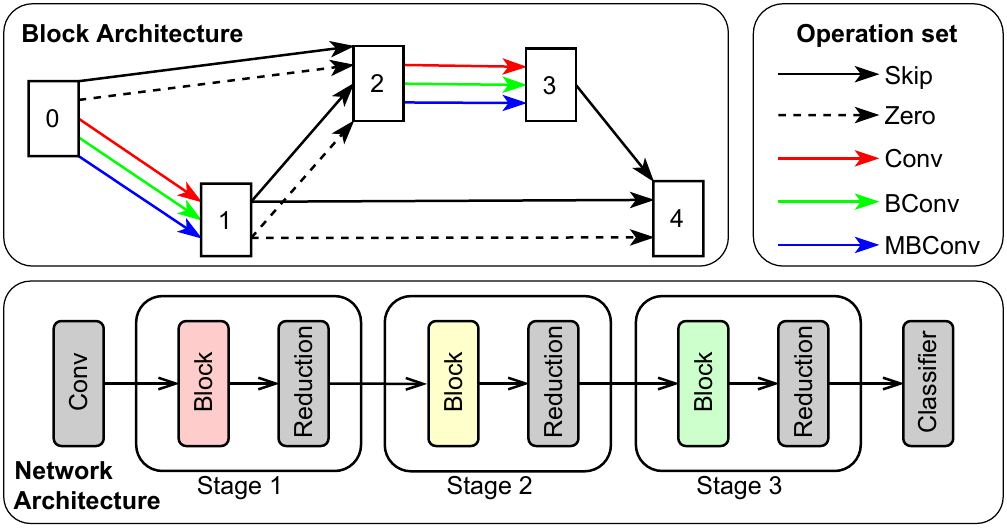}{trim=0 0 0 0}{Overview of the \blox{} macro search space.}{fig:blox_search_space}

\blox{} is a macro search space for CNNs on image classification task.
The search space is designed to be compatible with all NAS methods, including differentiable architecture search~\cite{darts}.

\subsection{Search space}

We opt for a simple search space definition that mimics many recent CNN architectures~\cite{mobilenetv3,efficientnetv1,efficientnetv2} and NAS search spaces~\cite{darts,nasbench2}.
Figure~\ref{fig:blox_search_space} shows an overview of the \blox{} search space.
The network architecture consists of three stages, each containing a searchable block and a fixed reduction block.
The searchable block can be expressed as a differentiable supernet as shown in the figure (block architecture), and is allowed to be different for each stage to construct the macro search space.
We designed the block architecture to allow for interesting and diverse connectivity between operations as exemplified in Figure~\ref{fig:blocks_examples}.
We make sure to include common structures such as residual connections~\cite{resnet} and inverted bottleneck blocks~\cite{efficientnetv2} that are relevant for the state-of-the-art CNNs.
In total, there are 45 unique blocks, making the size of the \blox{} search space $45^3=91,125$.
Additionally, we selected operations from the relevant state-of-the-art DNNs~\cite{mobilenetv3,efficientnetv2,resnet,vgg}, and controlled their repetition factor to roughly balance FLOPs and parameters across the different blocks (more details in the supplementary material).


\figvs{0.8}{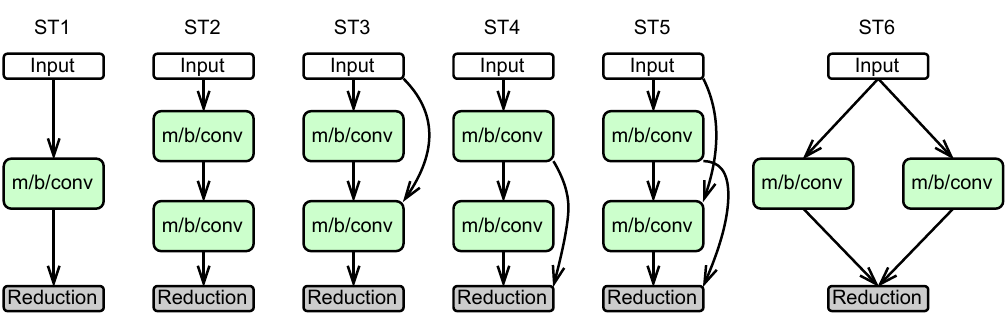}{trim=0 0 0 0}{Example block architectures from \blox{} showing diverse connectivities. Conv: VGG-style~\cite{vgg} 3x3 convolutions. BConv: Resnet-style~\cite{resnet} bottleneck with 5x5 depthwise-separable convolutions. MBConv: EfficientnetV2 fused-inverted residual convolution~\cite{efficientnetv2} including squeeze and excitation operation~\cite{squeeze-and-excitation}.}{fig:blocks_examples}



\subsection{Training details}

Throughout the paper we consider models from our \blox{} search space in 3 different training scenarios: \textit{1) Normal} setting is when a model is trained in a standard way, without any other model participating in the process. 
Information about the performance of all 91,125 models when trained normally comes pre-computed with our benchmark;
\textit{2) Distillation} refers to a setting in which individual candidate blocks are distilled independently to mimic analogous blocks from a normally-trained teacher model $T$ -- this process is described in more details in section~\ref{sec:algo}; 
\textit{3) Fine-tuning}, which follows \textit{Distillation}, is a process when blocks that were distilled independently are used to form an end-to-end model $M$. 
$M$ is then trained using the standard knowledge distillation approach with the same teacher $T$ which was used to distill blocks.
For the information about hyperparameters used for each of the three settings and what is included with the benchmark, please refer to the supplementary material.

\blox{} currently provides tabular results of training-from-scratch to enable systematic study of conventional NAS algorithms on emerging macro search spaces. Such results are also beneficial for studying blockwise algorithms (even though it does not directly enable their fast evaluation) because it allows better control of parameters of experiments (e.g. choosing "good teacher vs. bad teacher"), and enables comparison of the accuracy of the same models trained using different approaches.

\begin{table*}[!t]
\small
\centering
\caption {Comparison to other NAS benchmarks. }
\begin{tabular}{cccl}
\toprule
 & \# models & Type & Operations \\
\midrule
NAS-Bench-101~\cite{nasbench1} & 423k & \multirow{6}{*}{cell-based} & conv3x3, conv1x1, maxpool3x3 \\
NAS-Bench-201~\cite{nasbench2} & 15,625 &  & zeroize, skip connection, conv1x1, conv3x3,  \\
 & & & avgpool3x3 \\
NAS-Bench-1shot1~\cite{nasbench1shot1} & 363k & & conv3x3, conv1x1, maxpool3x3 \\ 
NAS-Bench-NLP~\cite{nasbenchnlp} & 14,322 & & linear, element wise, \\
 & & & activations (Tanh, Sigmoid, LeakyReLU) \\
NAS-Bench-ASR~\cite{nasbenchasr} & 8,242 & & linear, conv1x5, conv1x5 dilation2, \\
 & & & conv1x7, conv1x7 dilation2, zeroize \\
\midrule
NAS-Bench-Macro~\cite{nasbenchmacro} & 6,561 & \multirow{2}{*}{macro} & identity, MB3\_K3, MB6\_K6 \\
\blox{} & 91,125 & & conv, bconv, mbconv \\
\bottomrule
\end{tabular}
\label{tab:nasbenchmarks}
\end{table*}

\subsection{Differences to other NAS benchmarks}

\begin{figure}
\begin{minipage}[t]{0.55\linewidth}
    \centering
    \includegraphics[width=.9\textwidth]{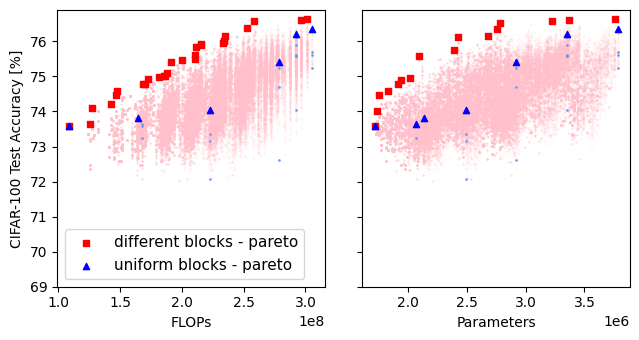}

    \caption{Accuracy v.s. FLOPs and parameters for all models in the \blox{} space. The Pareto-frontier of models with different blocks dominates that of models with repeated ``uniform" blocks -- only macro NAS can discover the superior models.}
    \label{fig:micro_vs_macro}
\end{minipage}
\hfill
\begin{minipage}[t]{0.43\linewidth}
    \centering
    \includegraphics[width=.9\linewidth]{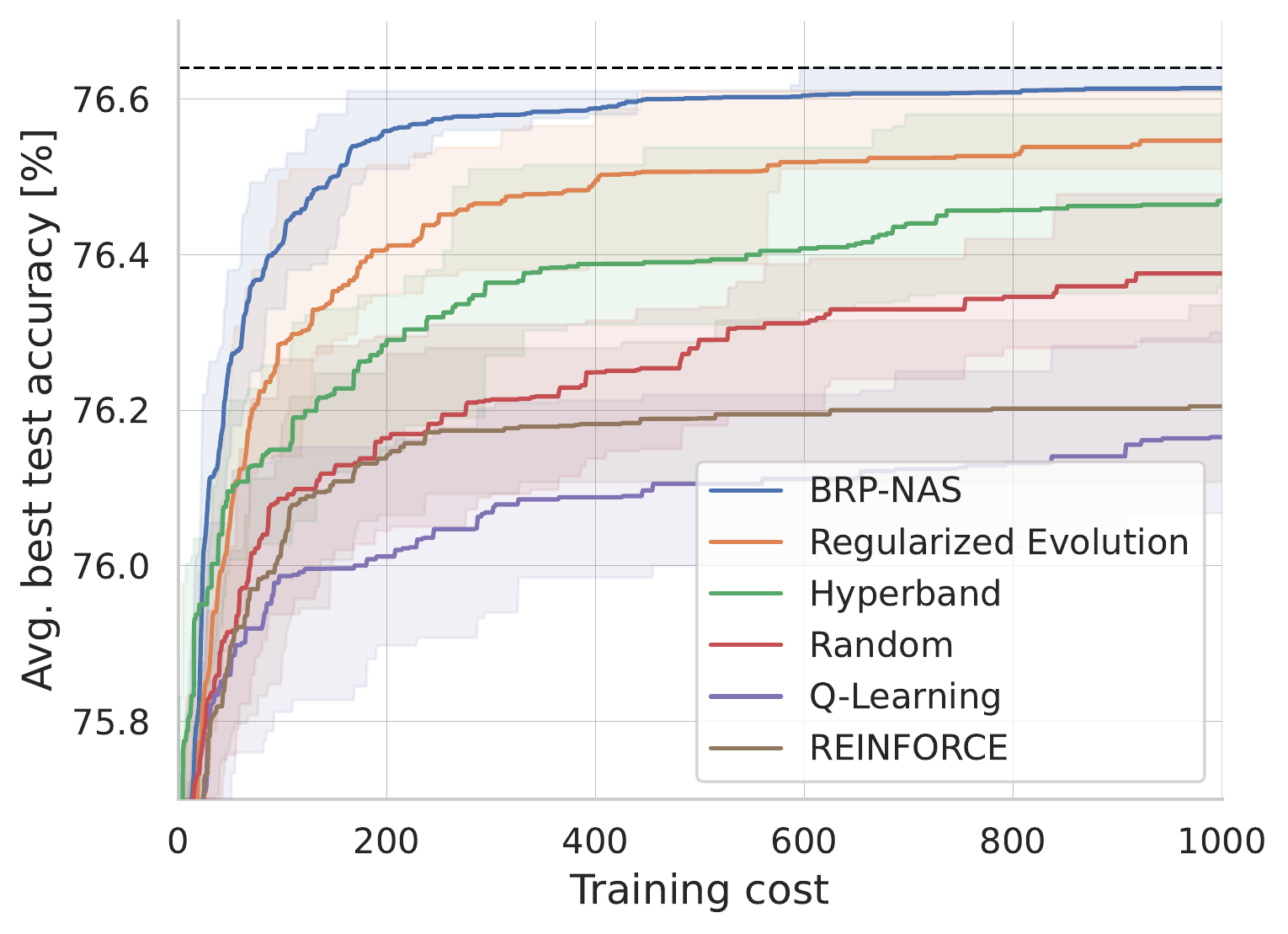}
    \caption{Comparison of conventional NAS search algorithms on \blox{}. For the details about each algorithm, please see Appendix~\ref{sec:nas_methods}.}
    \label{fig:conventional_nas}
\end{minipage}%
\end{figure}

We summarize characteristics of \blox{} and other recent NAS benchmarks in Table~\ref{tab:nasbenchmarks}.
In order to highlight both promises and challenges of macro NAS versus cell-based NAS, Figure~\ref{fig:micro_vs_macro} shows the accuracy of cell-based models consisting of uniform blocks (blue), and macro models consisting of different blocks (red), from our \blox{} search space when plotted against their number of FLOPs or parameters.
Pareto-optimal points are additionally emphasized with markers.
There are two highlights.
\textit{1)} The Pareto-frontier of models with different blocks clearly dominates that of models with uniform blocks.
It indicates that a macro search space contains higher performing models than a cell-based search space thus motivating our benchmark and study;
\textit{2)} There are many more models with different blocks than the models with uniform blocks.
The macro search space is much larger, posing a challenge to the searching algorithms.
Figure~\ref{fig:micro_vs_macro} shows the trade-off between achievable results and the amount of configurations available. Every cell-based search space can be turned into a much larger macro search space, which leads to a much higher exploration cost, and the achievable accuracy would likely improve.

Comparing to NAS-Bench-Macro, another published macro search space, \blox{} considers a larger number of diverse replacements. 
In terms of individual linear operations (e.g. a single convolution), the shallowest block out of the 45 candidates in \blox{} contains only 4 layers while the deepest block has 36 layers. 
This means that the depth of the whole network can range from 12 to 108 layers (excluding fixed parts). 
The design of \blox{} follows a complementary approach which uses a lower granularity of blocks with more diverse replacements, while NAS-Bench-Macro focuses on the opposite direction with higher granularity of blocks which results in lower diversity of candidates (e.g. NAS-Bench-Macro contains a single operation without any choices regarding connectivity, while \blox{} uses 2 operations per searchable stage thus introducing another degree of freedom related to the connections between them).
Having NAS benchmarks that explore different design choices increases our opportunities to study NAS algorithms in different situations and better understand their behaviour.

\section{Experiments: Characterizing Blockwise NAS on Blox}\label{sec:algo}
In order to alleviate the challenges associated with macro NAS, \textit{blockwise} algorithms have been proposed recently and showed promising results in optimising state-of-the-art models on large-scale image classification~\cite{dna2020,donna2021}.
Figure~\ref{fig:blockwise_flow_chart} shows an overview of blockwise NAS methods: \textit{1) Blockwise distillation} divides a pre-trained reference model (teacher) into sequential blocks that are later distilled into their possible replacements independently from each other.
The process of blockwise distillation produces a \textit{library} of pre-trained replacement blocks together with their \textit{signatures}, such as distillation loss or drop in the teacher's accuracy when a student block is swapped-in;
\textit{2) Search} uses these signatures to guide an algorithm to find well-performing models built by stacking a number of blocks from the block library;
\textit{3) Fine-tuning} is a process when blocks of student model are initialized with weights obtained in distillation, then the model is trained with knowledge distilled from the teacher.

\figvs{1.0}{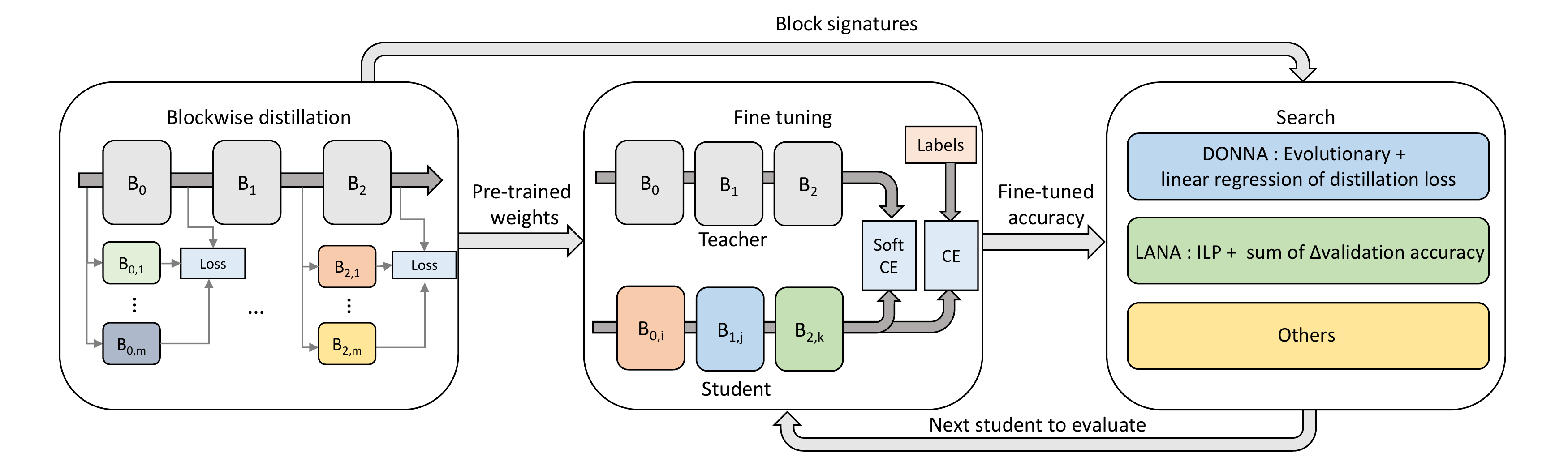}{trim=0 0 0 0}{Blockwise NAS -- (1) \textit{Blockwise distillation} is performed to obtain the signature of each candidate blocks. (2) \textit{Fine-tuning} initializes the blocks of student model with weights obtained in distillation. Then the student model is trained with knowledge distilled from the teacher. (3) \textit{Search} is conducted using different NAS algorithms to find the best model after fine-tuning.}{fig:blockwise_flow_chart}

Although outstanding results were demonstrated, blockwise NAS algorithms have not been thoroughly evaluated yet, presumably due to their exceptionally challenging setting.
To the best of our knowledge, their performance has not been evaluated in a common setting, nor compared to standard NAS methods, and very few of their design choices and assumptions have been adequately investigated. 
In this section, we attempt to fill these gaps with the help of our \blox{} search space and benchmark.

We begin by establishing a baseline by running conventional NAS algorithms that can be found in the literature in the simplest setting when each model is trained normally -- results are shown in Figure~\ref{fig:conventional_nas}. 
The relative efficiency of our search algorithms matches what is found in the literature.
Binary-relation predictor-based NAS (BRP-NAS)~\cite{brpnas2020} performs best, followed by evolutionary search~\cite{evonas} then other methods~\cite{hyperband2017,baker2017designing,rlnas2018}.
Other than providing these measurements to accompany our benchmark, we aim to compare to the two most recent blockwise NAS algorithms in the remainder of the paper --
\textbf{DONNA}~\cite{donna2021} employs a block-level knowledge distillation technique.
Each block's distillation loss is treated as its signature.
To perform search, an accuracy predictor (linear regression model) is trained by sampling and fine-tuning random architectures once all blocks are distilled.
The predictor takes the block signatures as input and predicts the performance of models.
This accuracy predictor guides an evolutionary search over the search space to find models that satisfy performance constraints.
\textbf{LANA}~\cite{hant} also uses blockwise distillation to train a library of blocks.
The block signature is the change of teacher's validation accuracy after a block is swapped with the candidate block.
Then an integer optimization problem, which minimize the sum of block signature, is used to select efficient models.




\subsection{Fine-tuning versus normal training}
We ask questions related to the performance of models when they are fine-tuned in the blockwise setting compared to that when trained normally, with special attention to the implications for NAS.

\begin{figure}
    \begin{minipage}[t]{0.45\textwidth}
        \centering
        \centering
            \includegraphics[width=\textwidth,trim={7inch 0.4inch 1inch 0.8inch},clip]{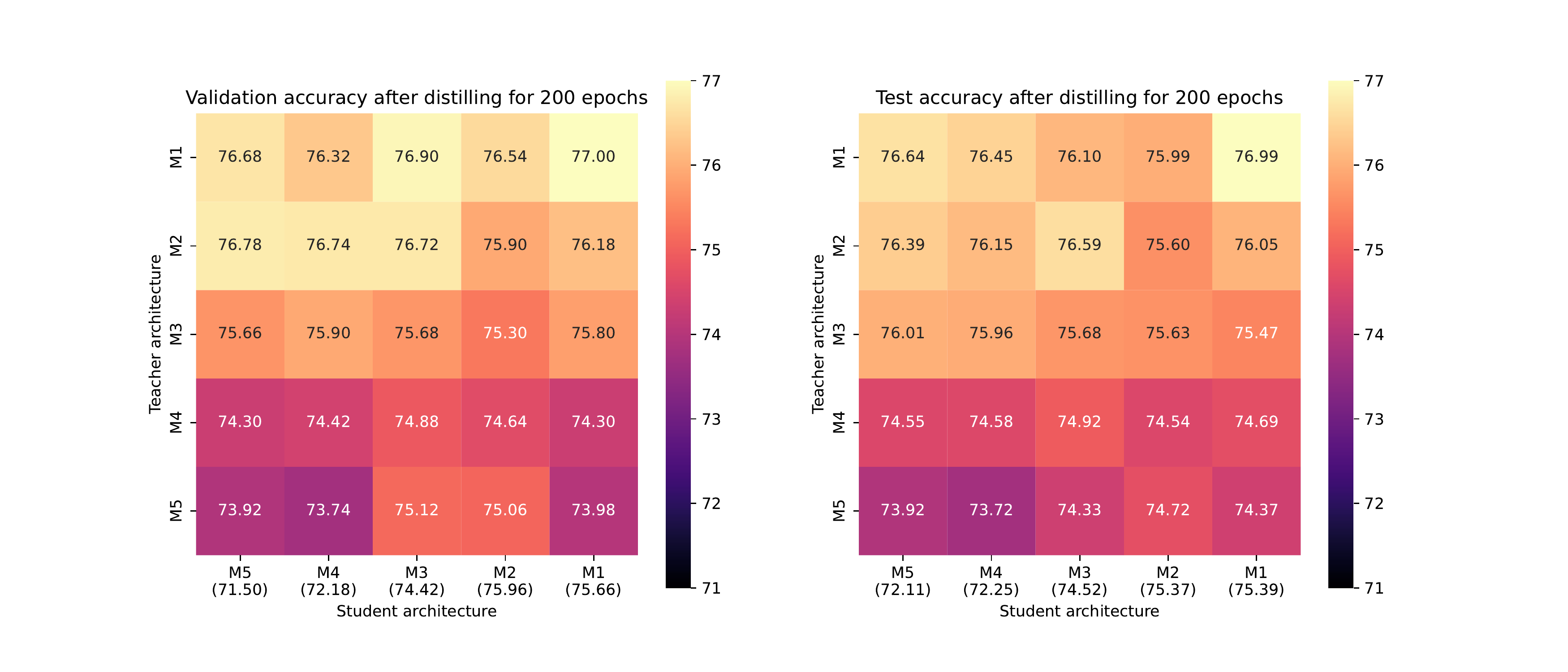}
            \caption{Mutual effect of the student and teacher architecture on the distillation outcome. Each cell at position $(x,y)$ contains information about accuracy when model M$x$ is blockwise-distilled and then fine-tuned from a normally-trained model M$y$. Performance of each model when trained normally is included in the X-axis' tick labels for reference.}
            \label{fig:teacher_vs_student}
     \end{minipage}
     \hfill
     \begin{minipage}[t]{0.50\textwidth}
         \centering
         \includegraphics[width=\textwidth]{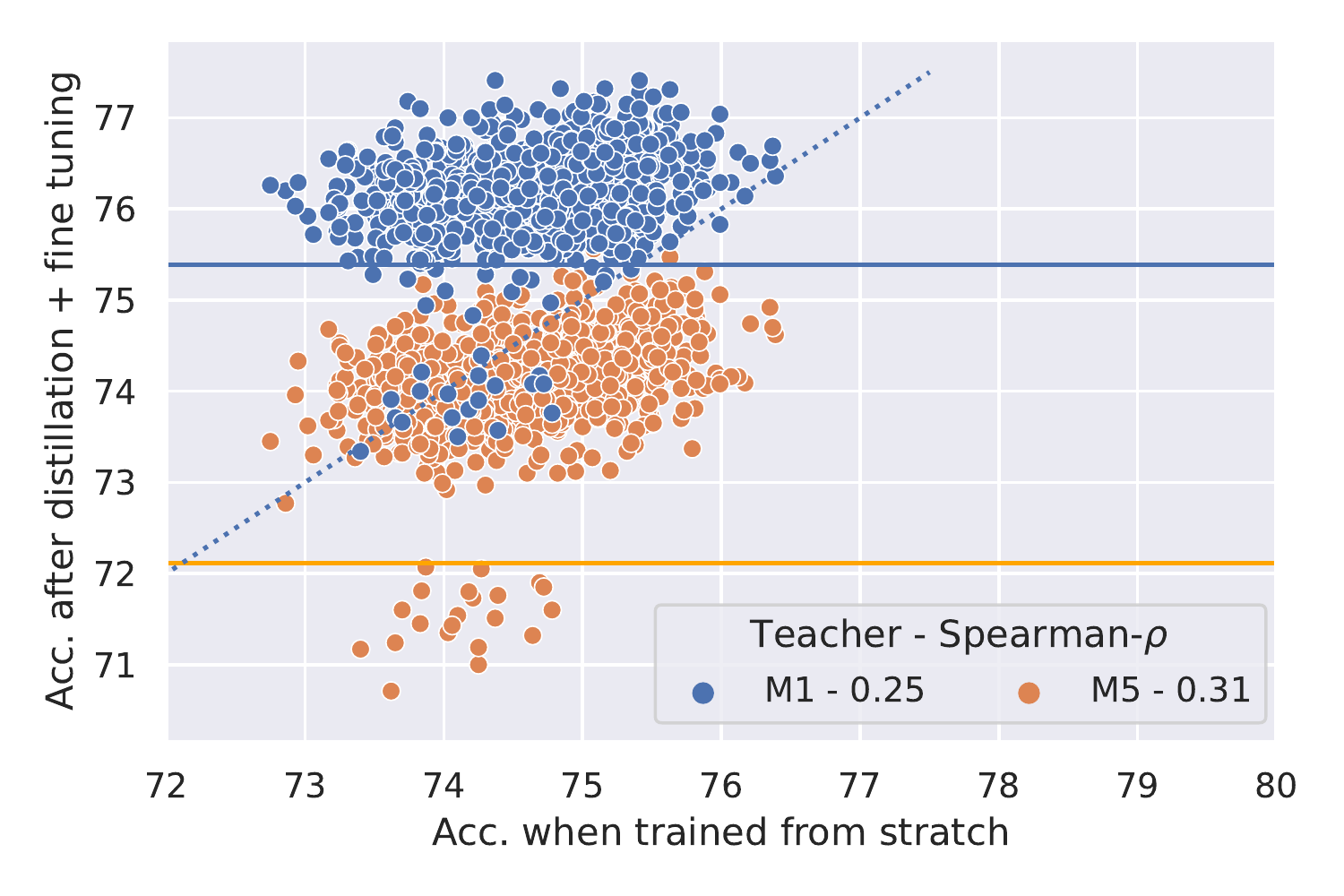}
         \caption{Spearman correlation between fine-tuned accuracy and training-from-scratch accuracy. For fine-tuning, all models are trained for 200 epochs, either using the good teacher (M1) or the bad teacher (M5). Solid lines mark performance of each teacher and dashed line marks $y=x$ diagonal.}
         \label{fig:ft_vs_gt_200}
     \end{minipage}
\end{figure}

\question{Does distillation help us achieve better performance compared to normal training?}
\label{q:dist_acc}
Distillation from a teacher model is a central part of both DONNA and LANA, at the same time there is plenty of evidence in the existing literature suggesting that distillation helps a model achieve better results, often exceeding even the teacher's performance.
However, it is important to note that the setting in those works is usually very different from NAS.
Specifically, distillation is conventionally used in situations when the teacher model is known to perform better than the student (e.g., it is significantly larger) -- in general, this important assumption might not hold in a NAS setting when we sample models from a search space without knowing if they are better or worse than our teacher.
In order to investigate the expected outcome of fine-tuning in different scenarios, we select 5 different architectures from our search space: 2 from the top performing ones, 1 average, and 2 bad ones; we refer to them as M1-M5, where M1 is the most accurate and M5 is the least accurate among them.
We then run blockwise distillation for 10 epochs and fine-tuning for 200 epochs (to match normal training) for each of the 25 possible $(student,teacher)$ pairs.
From the results in Figure~\ref{fig:teacher_vs_student}, we can see that in all cases \textbf{student model is able to improve upon its teacher, delivering on the promise of blockwise distillation from the existing works.}
However, we can also see that \textbf{compared to the accuracy achievable when a student is trained normally, fine-tuning does not always result in improvement.}
Specifically, when a bad teacher is used, accuracy of the models that otherwise tend to achieve good performance is suppressed -- this can be seen in the lower-right corner of Figure~\ref{fig:teacher_vs_student}.

\figvs{0.8}{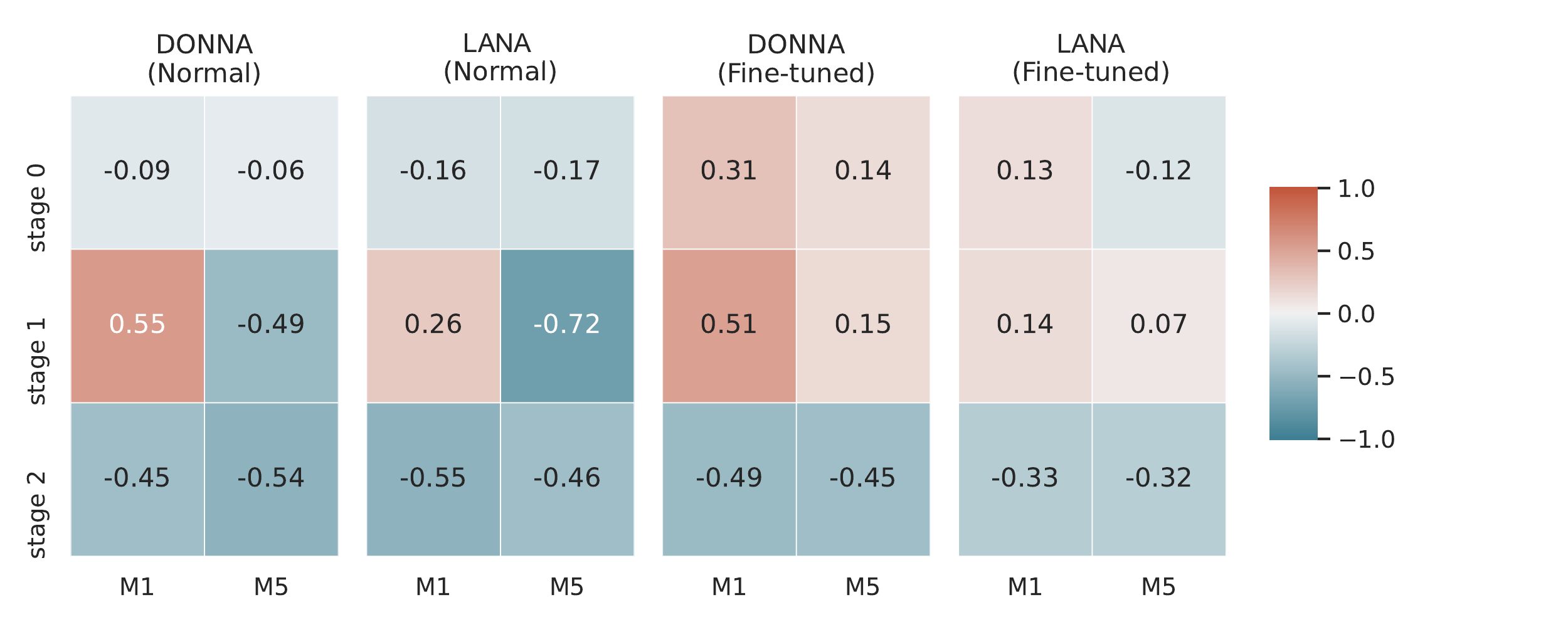}{trim=0 0 0 0}{Spearman correlation between block signatures and oracle ranking for 1000 random models. The oracle rank of a block is defined by the best normal / fine-tuned accuracy of a model containing that block. M1 and M5 are used as the teachers.}{fig:signature_vs_acc}

\question{Does fine-tuning accuracy correlate to training-from-scratch accuracy?}
\label{q:dist_normal_correlation}
Figure~\ref{fig:teacher_vs_student} includes one more significant observation for NAS -- even within our small sample of 5 models relative ranking of models after fine-tuning is different from when the models are trained normally.
For example, when M1 is used as a teacher we can see that the second best model turns out to be M5, which normally is the worst.
This suggests that \textbf{models exhibit vastly different performance when distilled compared to normal training.}
To further investigate this behaviour, we scale up our analysis to 1000 random student networks which are distilled with M1 and M5, then we correlate their training-from-scratch accuracy to fine-tuning accuracy.
Results are presented in Figure~\ref{fig:ft_vs_gt_200}.
We can see that indeed on the larger sample correlation remains rather weak for both teachers, with Spearman-$\rho$ of $0.25$ for M1 and $0.31$ for M5.
At the same time, Figure~\ref{fig:ft_vs_gt_200} further supports our observation that fine-tuning is only beneficial if a good teacher is used -- even though most of the students are able to significantly improve upon the M5 teacher, most of them do not improve upon their own training-from-scratch accuracy; this is not the case for a good teacher though.
Poor correlation between fine-tuning and normal training suggests that in general we should not perform NAS by simply searching for a good model using standard training and then rely on distillation to boost its performance -- instead, \textbf{we can achieve better results if we directly search for a model that performs well when distilled.}

\subsection{Searching for good students efficiently} \label{sec:efficient_search}

In the previous subsection we showed that blockwise distillation can be helpful in improving accuracy of models, and it is important to identify good students under distillation settings as fast as possible in order to minimize searching cost.
We also highlighted that blockwise methods utilise a divide-and-conquer approach where signatures of different blocks are used to guide the search.
We therefore ask the following questions related to block signatures and their usage in NAS.

\question{How well do block signatures identify good blocks?} 
\label{q:signatures_vs_oracle}
We compare different block scoring methods by measuring their correlation with an \textit{oracle ranking}.
The oracle ranking is computed by answering \textit{"if this block is selected at this stage, what is the best accuracy we can get?"} for each candidate block, and the blocks are then sorted accordingly.
Figure~\ref{fig:signature_vs_acc} compares the two block signatures, distillation loss (DONNA) and change of validation accuracy (LANA), in their ability to identify good blocks with approximated oracle ranking, when distilled from a good (M1) and a bad (M5) teacher.
It is the first time that the efficacy of DONNA and LANA block signatures are quantified, and surprisingly, \textbf{they are not consistently indicative of block performance.}
In particular, the correlation of the last stage (stage 2) is much worse than the earlier stages.



\question{Can we still use signatures to predict end-to-end performance?}
\label{q:prediction_from_signatures}
Even though signatures are not good indicators when it comes to identifying if individual blocks would lead to the best possible model on their own, it is still possible that they can be used in a smart way to estimate end-to-end performance.
In DONNA, a linear regression model with second-order terms is used to predict end-to-end accuracy, using block signatures as features and accuracy as targets.
In LANA, a simple sum of signatures is used as a proxy to approximate the non-linear objective to minimize the loss function.
At the same time, there are predictors that utilize graph structure rather than block signatures and deliver promising results in other settings.
For example, BRP-NAS~\cite{brpnas2020} uses a graph convolutional network (GCN) to capture graph topology and predict performance of a model.
We compare these different prediction-based approaches in estimating end-to-end model accuracy.

\figvs{1}{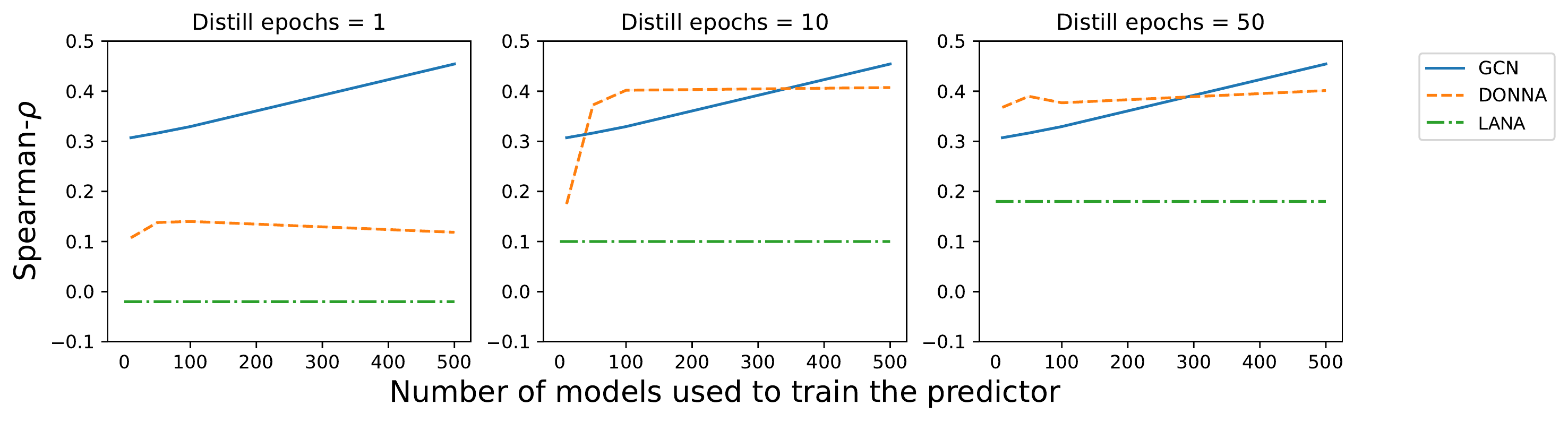}{trim=0 0 0 0}{Comparison of different predictors on estimating end-to-end model accuracy (after distillation and fine-tuning). Y-axis shows the Spearman correlation between the predicted and actual accuracy. In this experiment, 1000 models are randomly sampled from \blox{}. The number of models used to train the predictors are indicated in the x-axis, and the rest of the models are used for testing.}{fig:predictors}

Figure~\ref{fig:predictors} shows comparison of different predictors used to estimate end-to-end model accuracy after distillation and fine-tuning. 
There are 3 findings: \textbf{\textit{1)} Distillation loss + linear regression (DONNA) is better than change of validation accuracy + simple summation (LANA).
\textit{2)} Signatures of blocks that were distilled for more epochs tend to produce better predictors.
3) The GCN predictor, which does not require any distillation signatures, can outperform DONNA. However, DONNA works better with a small number of training points, provided that the blockwise distillation was performed with 10 or more epochs.}


\question{Do we have to fine-tune for 200 epochs?}
\label{q:fine_tuning_epochs}
\figvs{1}{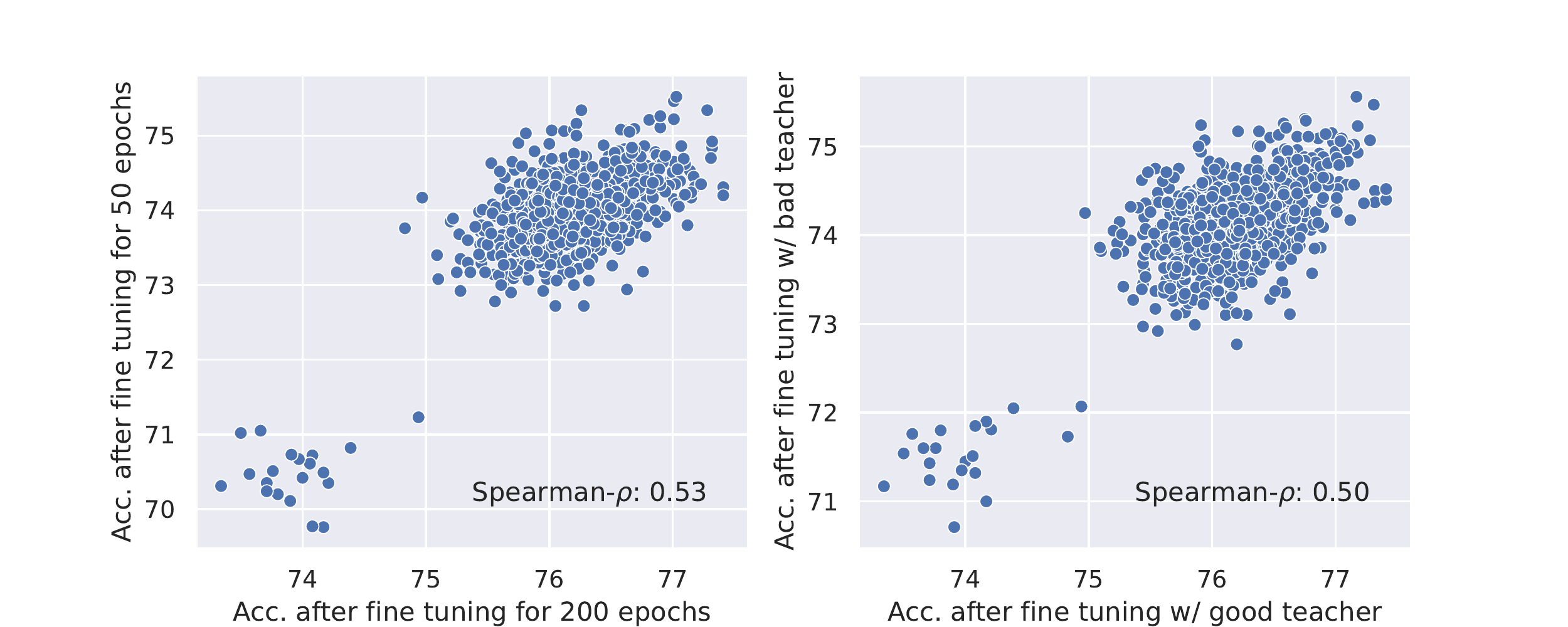}{trim=0 0 0 0}{(Left) Spearman correlation between fine-tuned accuracy for 50 epochs and 200 epochs, using the good teacher (M1). It indicates that high performing models can be identified by fine-tuning for fewer number of epochs. (Right) Spearman correlation between fine-tuned (for 200 epochs) accuracy using the good teacher (M1) and bad teacher (M5). It indicates that high performing models can be identified even by using the bad teacher.}{fig:ftacc_vs_epochs_and_teacher}

\begin{figure}[t]
    \begin{minipage}[t]{0.48\linewidth}
    \includegraphics[width=\linewidth]{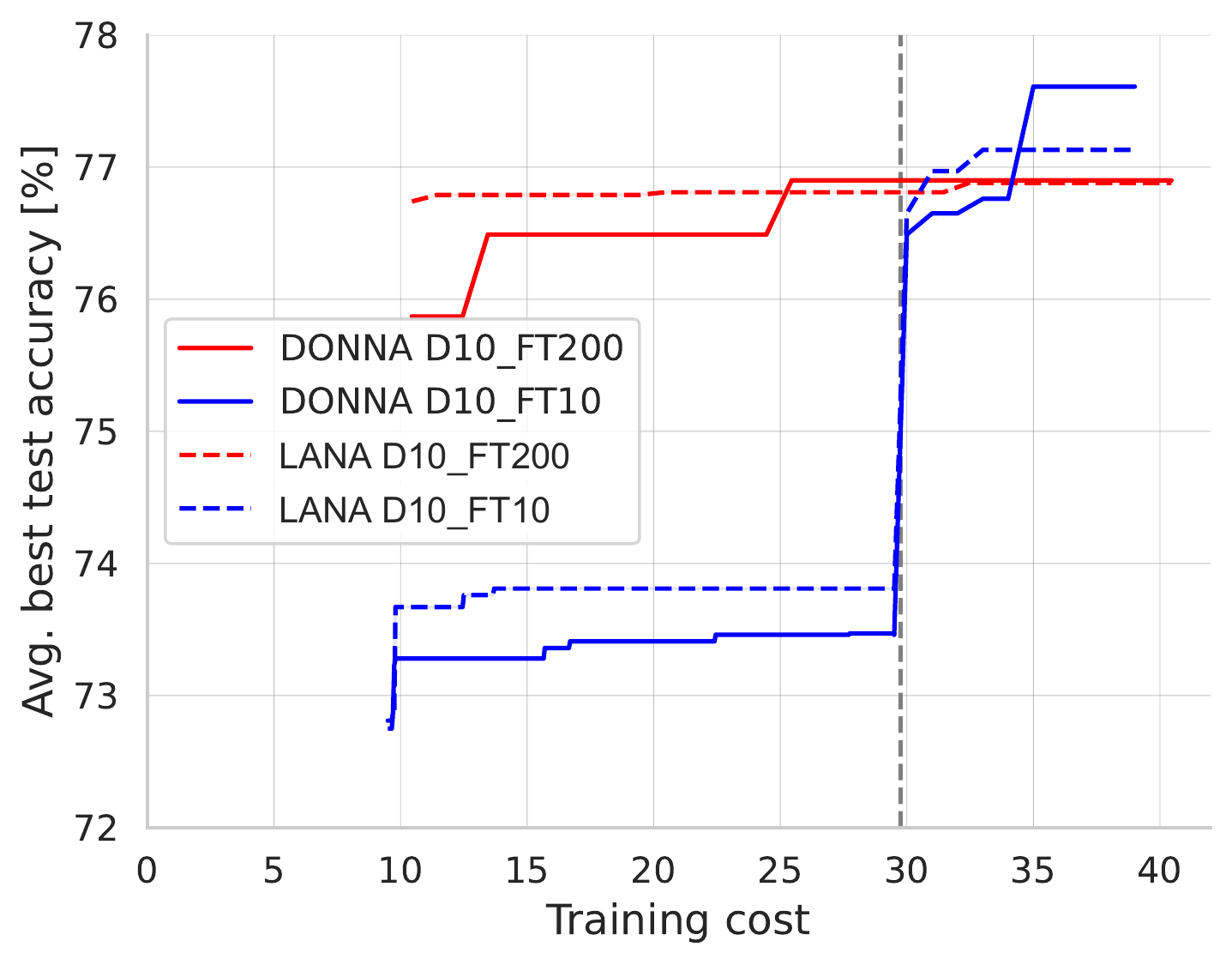}
    \caption{Comparison of models being searched and retrained using a different number of epochs. For the blue lines, the search is conducted until cost=30 with 10 fine-tuning epochs, after which the best models found are trained fully (to the right of the dashed gray line).}
    \label{fig:reduced_searching_epochs}
\end{minipage}%
\hfill
    \begin{minipage}[t]{0.48\linewidth}
    \includegraphics[width=\linewidth]{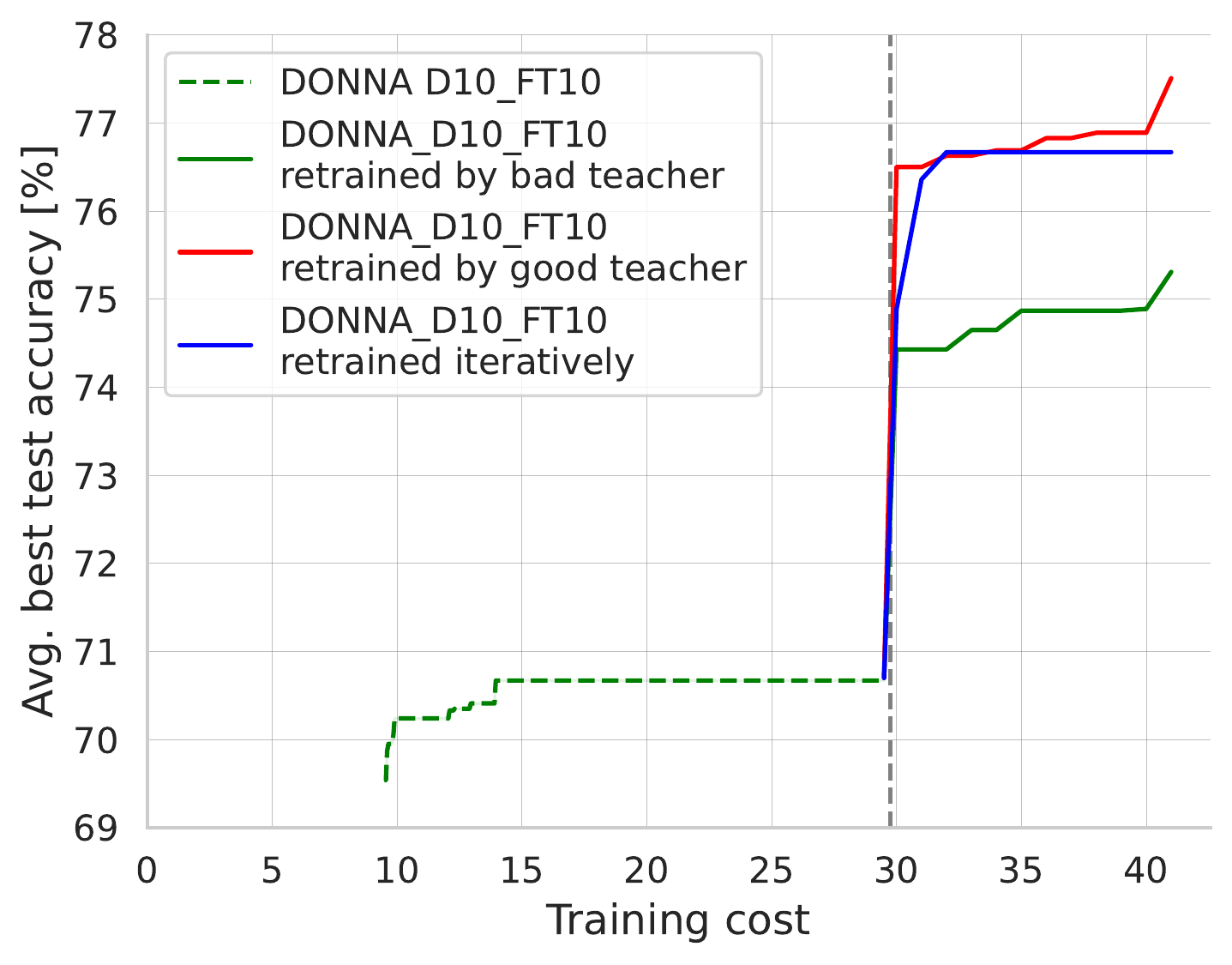}
    \caption{Models being searched using a bad teacher (M5), and retrained using different schemes. Firstly the search is conducted till cost=30 as indicated by the gray line. Then the models found during search are ranked and retrain for 200 epochs using 3 different schemes -- (1) the same bad teacher (green curve), (2) the good teacher (red curve), (3) iterative approach (blue).}
    \label{fig:varying_ft_teacher}
\end{minipage}%
\end{figure}

It is possible that reduced fine-tuning (e.g. for 50 epochs) can identify models that are as good as those found by the full searches (e.g. for 200 epochs).
To investigate this, Figure~\ref{fig:ftacc_vs_epochs_and_teacher} (left) shows the correlation of fine-tuning for 50 and 200 epochs -- the results suggest that \textbf{it should be possible to still identify good models without full fine-tuning.}
To confirm this hypothesis, in Figure~\ref{fig:reduced_searching_epochs}, we first search by fine-tuning the student models for 10 epochs (FT10, which has 400 models trained when the training cost reaches 30 as indicated by the gray line).
Then we rank the models searched and retrain them for 200 epochs using the same teacher.
The big improvement seen in the blue curves indicated that the models searched are good. 
After full training they outperform the models searched by fine-tuning the student models for 200 epochs (FT200, which only has 20 models searched when the training cost reaches 30).
This aligns with the results in Figure~\ref{fig:ftacc_vs_epochs_and_teacher} (left) that models trained with 50 epochs and 200 epochs are highly correlated.



\question{Can we search for good models without prior knowledge of a good teacher?}
\label{q:find_good_teacher}
Figure~\ref{fig:ftacc_vs_epochs_and_teacher} (right) shows the correlation of fine-tuning by different teachers.
We can see that the accuracy of models fine-tuned by a good teacher is highly correlated with that of a bad teacher.
Although we know that model selection is robust to the choice of teacher, it is not usually the case that we have prior knowledge of a good teacher.
In such case, we propose to find better teachers iteratively: 
1) Start NAS and fine-tune models using any teacher.
2) Stop the search and obtain a list of candidate models. 
3) Train the candidate models from scratch and pick the best model as the teacher.
4) Fine-tune the candidate models using the teacher selected, and pick the best model as the next teacher.
5) Repeat step 4 until we see no further improvement.
As we can see from the blue curve in Figure~\ref{fig:varying_ft_teacher}, \textbf{the iterative approach has significantly improved the model accuracy without knowing a good teacher in advance.}

\begin{table}
\caption {Performance of different NAS methods on \blox{} search space.
For blockwise-NAS methods, the blocks are distilled for 10 epochs and then the models are fine-tuned using a \textit{FT$\alpha$} setting, e.g. \textit{FT10} means the models are fine-tuned for 10 epochs in the search process.
In the cases that retraining is performed (which starts once the cost reaches 30), the searched models are ranked and retrained until the cost or accuracy target is reached, e.g. \textit{FT10 + FT200} means the models are retrained for 200 epochs.
If a different teacher is used for retraining, it is denoted, e.g. \textit{FT10 + FT200 \scriptsize{iter.}} means the teacher used in the retraining process is selected by the iterative strategy.
The middle column shows the accuracy achieved after reaching a fixed cost of 40, and the last column shows the cost required to achieve an accuracy of 76.6.
\XSolidBrush means the target accuracy cannot be achieved with reasonable cost.
$\dagger$~DARTS-PT is a differentiable NAS method that the cost of 40 does not apply.
}
\label{tab:summary}
\begin{minipage}[t]{.49\textwidth}
    \vspace{5mm}
    \centering
    \small
    \begin{tabular}[t]{lcc}
    \toprule
    \multirow{2}{*}{Method} & Acc. & Cost \\
    & @ cost=40 & @ acc.=76.6 \\
    \midrule
    \multicolumn{3}{l}{Conventional NAS} \\
    \cdashlinelr{1-3}
    BRP-NAS & 76.40 & 400 \\
    Reg. Evolution & 76.10 & \XSolidBrush \\
    DARTS-PT & 74.52$\dagger$ & \XSolidBrush \\ [0.2cm]
    \bottomrule
    \end{tabular}
\end{minipage}
\begin{minipage}[t]{0.49\textwidth}
    \vspace{5mm}
    \centering
    \small
    \begin{tabular}[t]{lcc}
    \toprule
    \multirow{2}{*}{Method} & Acc. & Cost \\
    & @ cost=40 & @ acc.=76.6 \\
    \midrule
    \multicolumn{3}{l}{Blockwise NAS \small{\textit{assuming good teacher (M1)}}} \\
    \cdashlinelr{1-3}
    FT200 & 76.90 &  25 \\
    FT10 & 73.47 & \XSolidBrush \\
    FT10 + FT200 & 77.66 & 30 \\[0.2cm]
    
    \multicolumn{3}{l}{Blockwise NAS \small{\textit{assuming bad teacher (M5)}}} \\
    \cdashlinelr{1-3}
    FT10 & 70.67 & \XSolidBrush \\
    FT10 + FT200 & 74.90 & \XSolidBrush \\
    FT10 + FT200 \scriptsize{iter.} & 76.67 & 30 \\
    FT10 + FT200 \scriptsize{M1} & 76.88 & 30 \\
    \bottomrule
    \end{tabular}
\end{minipage}
\end{table}

\subsection{Comparison of different NAS methods}

Table~\ref{tab:summary} quantifies the performance of different methods on the \blox{} search space.
We measure two things -- accuracy after reaching a fixed cost of 40, and cost required to achieve an accuracy of 76.6 (which is roughly the accuracy of the best model in our search space when trained normally).
For conventional NAS, we highlight \textit{regularized evolution}~\cite{evonas} and BRP-NAS~\cite{brpnas2020} which have the best results among the others, and \textit{DART-PT}~\cite{dartspt} which is a well-known differentiable NAS method.

We can see a few things: 
\textit{1)} Conventional NAS achieves worse results than standard blockwise (FT200) when a good teacher is used; 
\textit{2)} We can improve blockwise NAS by utilising reduced fine-tuning proxy followed by full fine-tuning (FT10 + FT200), which is our contribution stemming from questions Q~\ref{q:dist_acc}-\ref{q:fine_tuning_epochs}; 
\textit{3)} However, when a bad teacher is used (FT10 + FT200 at the bottom part), blockwise NAS actually falls short to its conventional counterpart -- the results can be improved by our proposed simple iterative strategy (FT10 + FT200 {\scriptsize{iter.}}, Q~\ref{q:find_good_teacher}), which allows us to again dominate conventional NAS. In fact, iterative strategy is almost as good as using the good teacher in the second phase of the search (FT10 + FT200 {\scriptsize{M1}}).

Overall, our results again showcase the dominant role of the final fine-tuning and, more broadly, quality of training in blockwise NAS.
We include more detailed discussion about interpretation of some of our results, as well as their limitations, in the supplementary material.


\section{Conclusion}\label{sec:conclusion}
In this work, we presented \blox{} -- a macro NAS search space and benchmark designed to provide a challenging setting for NAS.
With its help, we perform a thorough analysis of the emerging blockwise NAS algorithms and compare them to each other and the conventional NAS methods that can be found in the literature.
Our results include a quantitative analysis of the efficacy of block signatures and accuracy predictors, furthermore, we discover that the training methodology, especially the teacher model architecture during distillation, plays a bigger role than student model architecture in finding a good student model.
Our findings are somewhat unexpected and only made possible by the availability of \blox{}, for which we hope to see many more interesting studies by the research community.


\subsubsection*{Acknowledgments and disclosure of funding}
This work was done as a part of the authors’ jobs at the Samsung AI Center, and was supported by the National AI Strategy Fund at the Alan Turing Institute.
We thank the reviewers of the NeurIPS Datasets and Benchmarks Track 2022 for their comments and suggestions that helped improve the paper.

\clearpage
\bibliographystyle{unsrt}
\bibliography{references}

\begin{thebibliography}{10}

\bibitem{alexnet}
Alex Krizhevsky, Ilya Sutskever, and Geoffrey~E Hinton.
\newblock Imagenet classification with deep convolutional neural networks.
\newblock In {\em Advances in Neural Information Processing Systems}, 2012.

\bibitem{googlenet}
Christian Szegedy, Wei Liu, Yangqing Jia, Pierre Sermanet, Scott Reed, Dragomir
  Anguelov, Dumitru Erhan, Vincent Vanhoucke, and Andrew Rabinovich.
\newblock Going deeper with convolutions.
\newblock In {\em Proceedings of the IEEE/CVF Conference on Computer Vision and
  Pattern Recognition (CVPR)}, 2015.

\bibitem{mobilenetv3}
Andrew Howard, Mark Sandler, Bo~Chen, Weijun Wang, Liang-Chieh Chen, Mingxing
  Tan, Grace Chu, Vijay Vasudevan, Yukun Zhu, Ruoming Pang, Hartwig Adam, and
  Quoc Le.
\newblock Searching for mobilenetv3.
\newblock In {\em Proceedings of the IEEE/CVF International Conference on
  Computer Vision (ICCV)}, 2019.

\bibitem{efficientnetv1}
Mingxing Tan and Quoc Le.
\newblock {E}fficient{N}et: Rethinking model scaling for convolutional neural
  networks.
\newblock In {\em International Conference on Machine Learning (ICML)}, 2019.

\bibitem{efficientnetv2}
Mingxing Tan and Quoc~V. Le.
\newblock Efficientnetv2: Smaller models and faster training.
\newblock {\em CoRR}, abs/2104.00298, 2021.

\bibitem{rlnas}
Barret Zoph and Quoc~V. Le.
\newblock Neural architecture search with reinforcement learning.
\newblock In {\em International Conference on Learning Representations (ICLR)},
  2017.

\bibitem{evonas}
Esteban Real, Alok Aggarwal, Yanping Huang, and Quoc~V. Le.
\newblock Regularized evolution for image classifier architecture search.
\newblock {\em Proceedings of the AAAI Conference on Artificial Intelligence},
  33(01), Jul. 2019.

\bibitem{darts}
Hanxiao Liu, Karen Simonyan, and Yiming Yang.
\newblock {DARTS}: Differentiable architecture search.
\newblock In {\em International Conference on Learning Representations (ICLR)},
  2019.

\bibitem{zerocost}
Mohamed~S Abdelfattah, Abhinav Mehrotra, {\L}ukasz Dudziak, and Nicholas~Donald
  Lane.
\newblock Zero-cost proxies for lightweight nas.
\newblock In {\em International Conference on Learning Representations (ICLR)},
  2021.

\bibitem{dna2020}
Changlin Li, Jiefeng Peng, Liuchun Yuan, Guangrun Wang, Xiaodan Liang, Liang
  Lin, and Xiaojun Chang.
\newblock Blockwisely supervised neural architecture search with knowledge
  distillation.
\newblock In {\em Proceedings of the IEEE/CVF Conference on Computer Vision and
  Pattern Recognition (CVPR)}, 2020.

\bibitem{donna2021}
Bert Moons, Parham Noorzad, Andrii Skliar, Giovanni Mariani, Dushyant Mehta,
  Chris Lott, and Tijmen Blankevoort.
\newblock Distilling optimal neural networks: Rapid search in diverse spaces.
\newblock In {\em Proceedings of the IEEE/CVF International Conference on
  Computer Vision (ICCV)}, 2021.

\bibitem{hant}
Pavlo Molchanov, Jimmy Hall, Hongxu Yin, Jan Kautz, Nicol{\`{o}} Fusi, and
  Arash Vahdat.
\newblock {LANA:} latency aware network acceleration.
\newblock In {\em Proceedings of the European Conference on Computer Vision
  (ECCV)}, 2022.

\bibitem{nasbench1}
Chris Ying, Aaron Klein, Eric Christiansen, Esteban Real, Kevin Murphy, and
  Frank Hutter.
\newblock {NAS-Bench-101}: Towards reproducible neural architecture search.
\newblock In {\em International Conference on Machine Learning (ICML)}, 2019.

\bibitem{nasbench2}
Xuanyi Dong and Yi~Yang.
\newblock {NAS-Bench-201}: Extending the scope of reproducible neural
  architecture search.
\newblock In {\em International Conference on Learning Representations (ICLR)},
  2020.

\bibitem{nao2018}
Renqian Luo, Fei Tian, Tao Qin, Enhong Chen, and Tie-Yan Liu.
\newblock Neural architecture optimization.
\newblock In {\em Advances in Neural Information Processing Systems}, 2018.

\bibitem{npenas2020}
Chen Wei, Chuang Niu, Yiping Tang, and Jimin Liang.
\newblock {NPENAS:} neural predictor guided evolution for neural architecture
  search.
\newblock {\em CoRR}, abs/2003.12857, 2020.

\bibitem{bonas2020}
Han Shi, Renjie Pi, Hang Xu, Zhenguo Li, James Kwok, and Tong Zhang.
\newblock Bridging the gap between sample-based and one-shot neural
  architecture search with bonas.
\newblock In {\em Advances in Neural Information Processing Systems}, 2020.

\bibitem{bananas2020}
Colin White, Willie Neiswanger, and Yash Savani.
\newblock {BANANAS:} bayesian optimization with neural architectures for neural
  architecture search.
\newblock {\em CoRR}, abs/1910.11858, 2019.

\bibitem{mnasnet2019}
Mingxing Tan, Bo~Chen, Ruoming Pang, Vijay Vasudevan, Mark Sandler, Andrew
  Howard, and Quoc~V. Le.
\newblock Mnasnet: Platform-aware neural architecture search for mobile.
\newblock In {\em Proceedings of the IEEE/CVF Conference on Computer Vision and
  Pattern Recognition (CVPR)}, 2019.

\bibitem{nasbench1shot1}
A.~Zela, J.~Siems, and F.~Hutter.
\newblock Nas-bench-1shot1: Benchmarking and dissecting one-shot neural
  architecture search.
\newblock In {\em International Conference on Learning Representations (ICLR)},
  2020.

\bibitem{nasbenchnlp}
Nikita Klyuchnikov, Ilya Trofimov, Ekaterina Artemova, Mikhail Salnikov,
  Maxim~V. Fedorov, and Evgeny Burnaev.
\newblock Nas-bench-nlp: Neural architecture search benchmark for natural
  language processing.
\newblock {\em CoRR}, abs/2006.07116, 2020.

\bibitem{nasbenchasr}
Abhinav Mehrotra, Alberto Gil C.~P. Ramos, Sourav Bhattacharya, {\L}ukasz
  Dudziak, Ravichander Vipperla, Thomas Chau, Mohamed~S Abdelfattah, Samin
  Ishtiaq, and Nicholas~Donald Lane.
\newblock Nas-bench-asr: Reproducible neural architecture search for speech
  recognition.
\newblock In {\em International Conference on Learning Representations (ICLR)},
  2021.

\bibitem{nbmacro}
Xiu Su, Tao Huang, Yanxi Li, Shan You, Fei Wang, Chen Qian, Changshui Zhang,
  and Chang Xu.
\newblock Prioritized architecture sampling with monto-carlo tree search.
\newblock In {\em Proceedings of the IEEE/CVF Conference on Computer Vision and
  Pattern Recognition (CVPR)}, 2021.

\bibitem{cifar}
Alex Krizhevsky.
\newblock {Learning Multiple Layers of Features from Tiny Images}, 2009.

\bibitem{resnet}
Kaiming He, Xiangyu Zhang, Shaoqing Ren, and Jian Sun.
\newblock Deep residual learning for image recognition.
\newblock In {\em Proceedings of the IEEE/CVF Conference on Computer Vision and
  Pattern Recognition (CVPR)}, 2016.

\bibitem{vgg}
Karen Simonyan and Andrew Zisserman.
\newblock Very deep convolutional networks for large-scale image recognition.
\newblock In {\em International Conference on Learning Representations (ICLR)},
  2015.

\bibitem{squeeze-and-excitation}
Jie Hu, Li~Shen, and Gang Sun.
\newblock Squeeze-and-excitation networks.
\newblock In {\em Proceedings of the IEEE/CVF Conference on Computer Vision and
  Pattern Recognition (CVPR)}, 2018.

\bibitem{nasbenchmacro}
Xiu Su, Tao Huang, Yanxi Li, Shan You, Fei Wang, Chen Qian, Changshui Zhang,
  and Chang Xu.
\newblock Prioritized architecture sampling with monto-carlo tree search.
\newblock In {\em Proceedings of the IEEE/CVF Conference on Computer Vision and
  Pattern Recognition (CVPR)}, 2021.

\bibitem{brpnas2020}
Lukasz Dudziak, Thomas Chau, Mohamed Abdelfattah, Royson Lee, Hyeji Kim, and
  Nicholas Lane.
\newblock Brp-nas: Prediction-based nas using gcns.
\newblock In {\em Advances in Neural Information Processing Systems}, 2020.

\bibitem{hyperband2017}
Lisha Li, Kevin~G. Jamieson, Giulia DeSalvo, Afshin Rostamizadeh, and Ameet~S.
  Talwalkar.
\newblock Hyperband: A novel bandit-based approach to hyperparameter
  optimization.
\newblock {\em J. Mach. Learn. Res.}, 18:185:1--185:52, 2017.

\bibitem{baker2017designing}
Bowen Baker, Otkrist Gupta, Nikhil Naik, and Ramesh Raskar.
\newblock Designing neural network architectures using reinforcement learning.
\newblock In {\em International Conference on Learning Representations (ICLR)},
  2017.

\bibitem{rlnas2018}
Barret Zoph, Vijay Vasudevan, Jonathon Shlens, and Quoc~V. Le.
\newblock Learning transferable architectures for scalable image recognition.
\newblock In {\em Proceedings of the IEEE/CVF Conference on Computer Vision and
  Pattern Recognition (CVPR)}, 2018.

\bibitem{dartspt}
Ruochen Wang, Minhao Cheng, Xiangning Chen, Xiaocheng Tang, and Cho-Jui Hsieh.
\newblock Rethinking architecture selection in differentiable nas.
\newblock In {\em International Conference on Learning Representations (ICLR)},
  2021.

\bibitem{ruder2016overview}
Sebastian Ruder.
\newblock An overview of gradient descent optimization algorithms.
\newblock {\em arXiv preprint arXiv:1609.04747}, 2016.

\bibitem{adam}
Diederik~P. Kingma and Jimmy Ba.
\newblock Adam: {A} method for stochastic optimization.
\newblock In {\em International Conference on Learning Representations (ICLR)},
  2015.

\bibitem{kd2015}
Geoffrey Hinton, Oriol Vinyals, and Jeffrey Dean.
\newblock Distilling the knowledge in a neural network.
\newblock In {\em NIPS Deep Learning and Representation Learning Workshop},
  2015.

\bibitem{cosine_annealing}
Ilya Loshchilov and Frank Hutter.
\newblock {SGDR:} stochastic gradient descent with warm restarts.
\newblock In {\em International Conference on Learning Representations (ICLR)},
  2017.

\bibitem{cubuk2019randaugment}
Ekin~D. Cubuk, Barret Zoph, Jonathon Shlens, and Quoc~V. Le.
\newblock Randaugment: Practical automated data augmentation with a reduced
  search space.
\newblock In {\em Proceedings of the IEEE/CVF Conference on Computer Vision and
  Pattern Recognition (CVPR)}, 2019.

\bibitem{playing_atari}
Volodymyr Mnih, Koray Kavukcuoglu, David Silver, Alex Graves, Ioannis
  Antonoglou, Daan Wierstra, and Martin Riedmiller.
\newblock Playing atari with deep reinforcement learning.
\newblock In {\em NIPS Deep Learning Workshop}, 2013.

\bibitem{reinforce}
R.~J. Williams.
\newblock Simple statistical gradient-following algorithms for connectionist
  reinforcement learning.
\newblock {\em Machine Learning}, 8:229--256, 1992.

\bibitem{datasheet_dataset}
Timnit Gebru, Jamie Morgenstern, Briana Vecchione, Jennifer~Wortman Vaughan,
  Hanna Wallach, Hal Daumé, and Kate Crawford.
\newblock Datasheets for datasets.
\newblock {\em arXiv preprint arXiv:1803.09010}, 2018.

\end{thebibliography}


\clearpage
\appendix
\addcontentsline{toc}{section}{Supplementary Material} 
\part{Supplementary Material} 
\parttoc 


\section{Access to the benchmark}
The source code and dataset are hosted under this URL :
{\small\url{https://github.com/SamsungLabs/blox}}.

The source code and dataset are licensed by the CC BY-NC ({\small\url{https://creativecommons.org/licenses/by-nc/4.0/}}) license. 
On the above url, this license is described as follows:

\begin{enumerate}
\item Copy and redistribute the material in any medium or format.
\item Remix, transform, and build upon the material for non-commercial purposes.
\item The licensor cannot revoke these freedoms as long as you follow the license terms.
\end{enumerate}

\section{Dataset generation}

Details to produce the dataset and experimental results are described below.
The instructions, source code and dataset can be found in the official repository.

\subsection{Benchmark metrics and training environment}

Table~\ref{tab:blox_statistics} summarizes the metrics and environment included in the \blox{} benchmark.
All the metrics can be queried via the API provided with the code.
In the dataset, we include the information obtained by \textit{Normal} training -- 
training and validation metrics (loss and accuracy) are logged at each epoch.
Test metrics and training time are logged at the end of training for the best model.
We also save the static information consisting of FLOPs, the number of parameters and an example architecture vector related with the entry (note that more than one architectural vector might map to the same architecture, therefore all entries across all parts of \blox{} are stored using hashes of graphs representing each model, similar to~\cite{nasbench1}).
Benchmarking information for each model are included as separate sections -- one file per benchmarking setting (device, input size, etc.).
In addition, information about the training environment are included for reproducibility.
That includes random seed, GPU used, versions of the OS, codebase, driver, etc.).
The dataset is fully modular, meaning that different sections can be used alone or combined with others freely.

\begin{table}
\begin{minipage}[t]{.45\textwidth}
    \small
    \caption {Information included in each part of the \blox{} dataset. Each section contains exactly 91,125 entries.}
    \label{tab:blox_statistics}
    \begin{tabularx}{\textwidth}{p{2cm} X}
    \toprule
    Section & Information \\
    \midrule
    \multirow{3}{*}{
        \shortstack[l]{Basic\\
        {\scriptsize \textit{(one per tr. seed)}}}
    }                               & Top1 validation accuracy \\
                                    & Top1 test accuracy \\
                                    & Training time \\
    \midrule
    \multirow{7}{*}{
        \shortstack[l]{Extended\\
        {\scriptsize \textit{(one per tr. seed)}}}
    }                               & Top1 training accuracy \\
                                    & Top5 training accuracy \\
                                    & Top5 validation accuracy \\
                                    & Top5 test accuracy \\
                                    & Training loss \\
                                    & Validation loss \\
                                    & Test loss \\
    \midrule
    \multirow{3}{*}{Static}         & FLOPs \\
                                    & Number of parameters \\
                                    & Architecture vector \\
    \midrule
    Benchmark                       & \multirow{2}{*}{Latency} \\
    {\scriptsize \textit{(one per device)}} & \\
    \midrule
    \multirow{5}{*}{
        \shortstack[l]{Environment\\
        {\scriptsize \textit{(one per tr. seed)}}}
    }                               & Python version \\
                                    & GPU model \\
                                    & Driver version \\
                                    & PyTorch version \\
                                    & Codebase version \\
    \bottomrule
    \end{tabularx}
\end{minipage}
\hspace{5mm}
\begin{minipage}[t]{.45\textwidth}
    \small
    \raggedright
    \caption{Training hyperparameters used throughout this work.}
    \label{tab:training_hyperm}
    \begin{tabular}{cccc}
    \toprule
    & Normal & Distillation & Fine-tuning \\ 
    \midrule
    Optimizer & SGD & Adam & Adam\\
    Loss & CE & MSE & KD ($\alpha$ = 0.9) \\
    T & N/A & N/A & 4 \\
    Epochs & 200 & [1,10,50] & [10,50,200] \\
    Momentum & 0.9 & N/A & N/A \\
    Betas & N/A & (0.9, 0.999) & (0.9, 0.999) \\
    LR schedule & cosine & cosine & cosine \\
    Initial LR & 0.01 & 0.005 & 5e-5 \\
    Final LR & 0 & 0 & 0 \\
    Weight decay & 0.0005 & 0.0005 & 0.0005\\
    Batch size & 256 & 256 & 256\\
    H. flip & 0.5 & \XSolidBrush & 0.5 \\
    Cutout & 1/16/1.0 & \XSolidBrush & 1/16/1.0 \\
    RandAug & 14/2 & \XSolidBrush & 14/2 \\
    Pad \& crop & 4 & \XSolidBrush & 4 \\
    \bottomrule
    \end{tabular}
\end{minipage}
\end{table}

\subsection{Training details}

Table~\ref{tab:training_hyperm} summarizes the hyperparameters used for each of the three training settings considered in our paper -- all hyperparameters were decided by performing a grid search using 5 random models from our search space (in case of distillation and fine-tuning the same model was used as both teacher and student).
All training has been done using PyTorch-based code on a single GPU (one of NVIDIA 1080Ti, 2080Ti, V100, P40, P100, and A40).
We train each architecture with the same strategy.
Specifically, we train each architecture via SGD~\cite{ruder2016overview} or Adam~\cite{adam}, using the cross-entropy (CE) loss, mean squared error (MSE) or knowledge distillation (KD)~\cite{kd2015} for 200 epochs in total. 
We set the weight decay as 0.0005 and decay the learning rate from 0.01/0.005/5e-5 to 0 with cosine annealing~\cite{cosine_annealing}.
For data augmentation, we use random horizontal flip with the probability of 0.5, random crop of 32×32 patch with 4 pixels padding on each border, cutout with 1.0 probability to cut one 16x16 patch out of each image.
We also use the RandAug scheme implemented by~\cite{cubuk2019randaugment}, 2 augmentation transformations are applied sequentially with the magnitude of 14.


\section{Implementation details}

\subsection{\blox{} search space}
\label{app:sec:arch_encoding}

\noindent \textbf{Operations.}
In the main paper (Section~\ref{sec:blox}), we presented several convolutional operations with residuals that are based on those in common networks~\cite{vgg,resnet,mobilenetv3,efficientnetv2}.
Here we gives more details on each of them.

\begin{itemize}
    \item Standard convolution (Conv) block - 3x3 convolution operation is repeated for 4 times.
    \item Bottleneck convolution (BConv) block - 
    This block consists of 6 repeated operations in the form of convolution -> bottleneck -> convolution.
    The first convolution is pointwise (1x1 convolution) with $N_i$ input channels and $N_i / b_i$ output channels, where $\{b_0, b_1, b_2, b_3, b_4, b_5\}$ = $\{2, 1, 0.5, 2, 1, 0.5\}$.
    Then a bottleneck operation (5x5 depthwise-separable convolution) is performed.
    Lastly, another pointwise convolution brings the representation back to $N_i$ output channels.
    Residual connection is applied to add input to the output.
    \item Inverted residual convolution (MBConv) - 
    This block consists of 2 repeated operations that follow a convolution -> squeeze-and-excitation -> convolution structure.
    $N_i$ input channels are first widen with an expansion ratio $r$ via 3x3 convolution, followed by a squeeze-and-excitation operation~\cite{squeeze-and-excitation}.
    Then a pointwise convolution reduces the number of output channels back to $N_i$.
    Finally residual connection is applied to add input to the output.
\end{itemize}

\begin{wrapfigure}[18]{r}{0.3\textwidth}
  \begin{center}
    \vspace{-5mm}
    \includegraphics[width=0.2\textwidth]{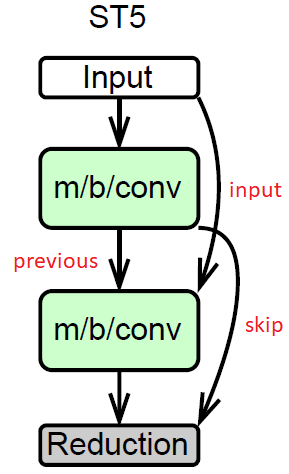}
  \end{center}
  \caption{ST5 block architecture is an example with all \textit{previous}, \textit{input} and \textit{skip} connections enabled.}
  \label{fig:st5_example}
\end{wrapfigure}

We control the repetition factor of each block to roughly balance FLOPs and parameters across the different blocks.
This is in order to avoid a situation when a certain operations is naturally a better choice not because of its structure and/or ability to efficiently use its parameters, but because it's significantly larger/smaller than other choices.
For example, compared to the standard convolution block, an inverted residual convolution contains significantly more parameters due to its expansion ratio -- to compensate for that, we repeat it fewer time.
Analogously, the bottleneck convolution block was originally proposed as a lightweight alternative to standard convolutions, therefore it naturally has significantly fewer parameters and FLOPS and we can afford to have more of them stacked together.
Although we tweak repetitions and other parameters (e.g., kernel size, expansion ratio, etc.) to minimize the differences in high-level metrics, please note that in general the differences are still there, just less significant than they could have been without this balancing.

\noindent \textbf{Architectures.}
There are 6 types of block architectures as shown in the main paper (Figure~\ref{fig:blocks_examples}).
Each block has two cells (except ST1 which has only one cell), each of which consists of 3 possible operations.
If we identify isomorphic cells in ST6 as well, there are 45 unique blocks, which account to $45^3$ = 91,125 total unique models in the search space.

In \blox{}, we encode each model architecture by a 15-dimensional architecture vector. 
An architecture vector is formatted as [$S_0$, $S_1$, $S_2$], which describes the 3 stages of searchable blocks.
Stage $S_i$ is described as [[$O_0$], [$O_1$, \textit{previous}, \textit{input}, \textit{skip}]], where $O_0$ and $O_1$ refer to the operations of the first and second cell, respectively.
\textit{previous} and \textit{input} are binary values that indicate if the cell has connections with the previous cell and the input of the block.
\textit{skip} is another binary value that indicates if there is any skip connection in the block.
Since $O_0$ only has one fixed connection to the input, no connection needs to be specified.
Figure~\ref{fig:st5_example} is an example of block architecture which has all connections enabled.

\subsection{NAS algorithms}
\label{sec:nas_methods}

Below we give the details of the conventional NAS algorithms used in the paper -- all algorithms operate on discreet architecture encodings as defined in Section~\ref{app:sec:arch_encoding}.
We did not perform any hyperparameter tuning for any of the algorithms - all our choices come from either the original papers that proposed relevant methods, or follow common practice that can be found in many NAS works.

All algorithms are optimizing for the best validation accuracy achieved by a model throughout its training -- if an algorithm alters the number of training epochs, this is the best validation accuracy for the reduced training.
When reporting test accuracy we assume each architecture has been trained fully.
Since our encoding allows for constructing invalid models (input and output of a single cell might be disconnected if $previous=input=skip=False$), we reward such models with a constant value of $-1$, these models do not contribute to the cost of running a search.

\noindent \textbf{Random search.}
We samples architectures from a uniform distribution over all possible architecture encodings.

\noindent \textbf{Q-Learning.}
We train a 4-layer MLP with 128 neurons in each layer to predict the total return of a two-step alternation process to a model's structure.
Given an initial architecture encoding (state), the agent can take an action that changes any single component of the encoding to a different value.
After two consecutive actions, the accuracy of the resulting model is considered to be the final reward and the resulting trajectory of 3 models is added to the training set of the agent.
We keep up to 32 last trajectories to enable history replay~\cite{playing_atari}, out of which we sample 8 to construct a single training batch for the agent.
Additionally, when deciding on an action to take, we always reject choices that would result in the model's structure reverting to any of the previous configurations within the same trajectory -- in other words, we reject any trajectories that would contain a cycle, e.g., $A \rightarrow B \rightarrow A$.
We use $\epsilon$-greedy strategy with $\epsilon=0.1$, and discounting with $\gamma=0.1$.
Adam~\cite{adam} optimizer is used to train the MLP, with learning rate set to $3.5\times 10^{-4}$, weight decay set to $0$, betas set to $(0.9, 0.999)$ and epsilon set to $1.0\times 10^{-8}$.
The reward for each valid model (validation accuracy) is normalized linearly to the range $[ 0, 1 ]$ using 0 and 100 as min and max, respectively.

\noindent \textbf{REINFORCE.}
We use a single cell LSTM controller trained with REINFORCE, following the setup in~\cite{rlnas2018}.
Specifically, the LSTM has 100 hidden and input features and is used auto-regressively to produce categorical distributions for each dimension of the architectural encoding.
The distributions is constructed using tanh activation with temperature $T=5$ and a constant scaling factor of $2.5$.
The initial input to the LSTM cell is all zeros.
The produced distributions are then sampled to decide on a model to train and the controller is then updated by using REINFORCE~\cite{reinforce} algorithm with Adam~\cite{adam} optimizer.
The optimizer's setting follows those described in the paragraph about our Q-Learning algorithm, we also use entropy-based regularization with weight $0.0001$.
Similar to Q-Learning, rewards for valid models are scaled linearly to the range $[0,1]$ using 0 and 100 as min and max, respectively.

\noindent \textbf{Hyperband.}
We use our implementation of Hyperband which follows description from the original paper~\cite{hyperband2017}.
We use $R=100$ and $\eta=3$.
The first value represents training budget in percentages, where $100$ corresponds to full training (200 epochs) and lower values scale the number of training epochs accordingly.
The choice of $\eta$ follows the recommendation in~\cite{hyperband2017}.
The hyperparameters exposed to the algorithm are the values constituting architecture encoding explained in~\ref{app:sec:arch_encoding}; we use uniform sampling to get random points.

\noindent \textbf{Regularized evolution.}
We run regularized evolution~\cite{evonas} with pool size 64 and sample size 16.

\noindent \textbf{BRP-NAS.}
We follow the implementation of BRP-NAS~\cite{brpnas2020} and use a 4-layer binary GCN predictor with 600 hidden units in each layer.
We use the official code available at \url{https://github.com/SamsungLabs/eagle} to perform training.
Specifically, we use the following hyperparameters: AdamW optimizer with LR set to $3.5\times 10^{-4}$ and weight decay set to $5.0\times 10^{-4}$, dropout rate of $0.2$, LR patience of 10, early stopping patience of 35, 250 training epochs, batch size of 64, $\alpha$ of 0.5.
We keep training the predictor every 20 models up to 200.
The input to the predictor is a graph representation of a model analogous to the ones shown in Figure~\ref{best_models} and~\ref{worst_models}, after operations are encoded as one-hot vectors and a global node is added, as explained in~\cite{brpnas2020}.
The same operations at different stages of a network are assigned different labels.

\subsection{Latency measurements}

We run each model in \blox{} on the follow devices (more devices will be added in the future). 
(i) Desktop platform - NVIDIA GTX 1080 Ti, 
(ii) Mobile platform - Qualcomm Snapdragon 888 with Hexagon 780 DSP.

We run each model 1000 times on each aforementioned device using a patch size of 32 x 32 and a batch size of 1 for mobile devices and 256 for desktop GPU.
For mobile devices, each model is quantized to 8 bits and run 10 time with the same settings using tools provided by Snapdragon Neural Processing Engine.
For desktop GPU, we use PyTorch and run each model after it is optimized and compiled using Torchscript.
In order to lessen the impact of any startup or cool-down effects such as the creation and loading of inputs into buffer, we discard latencies that fall outside the lower and higher quartile values before taking the average of every 10 runs. These averages are discarded again with the aforementioned thresholds before a final average is taken.

\section{Additional experimental results}

\subsection{Statistics of the \blox{} benchmark}

\begin{figure}
    \centering
    \includegraphics[width=\textwidth]{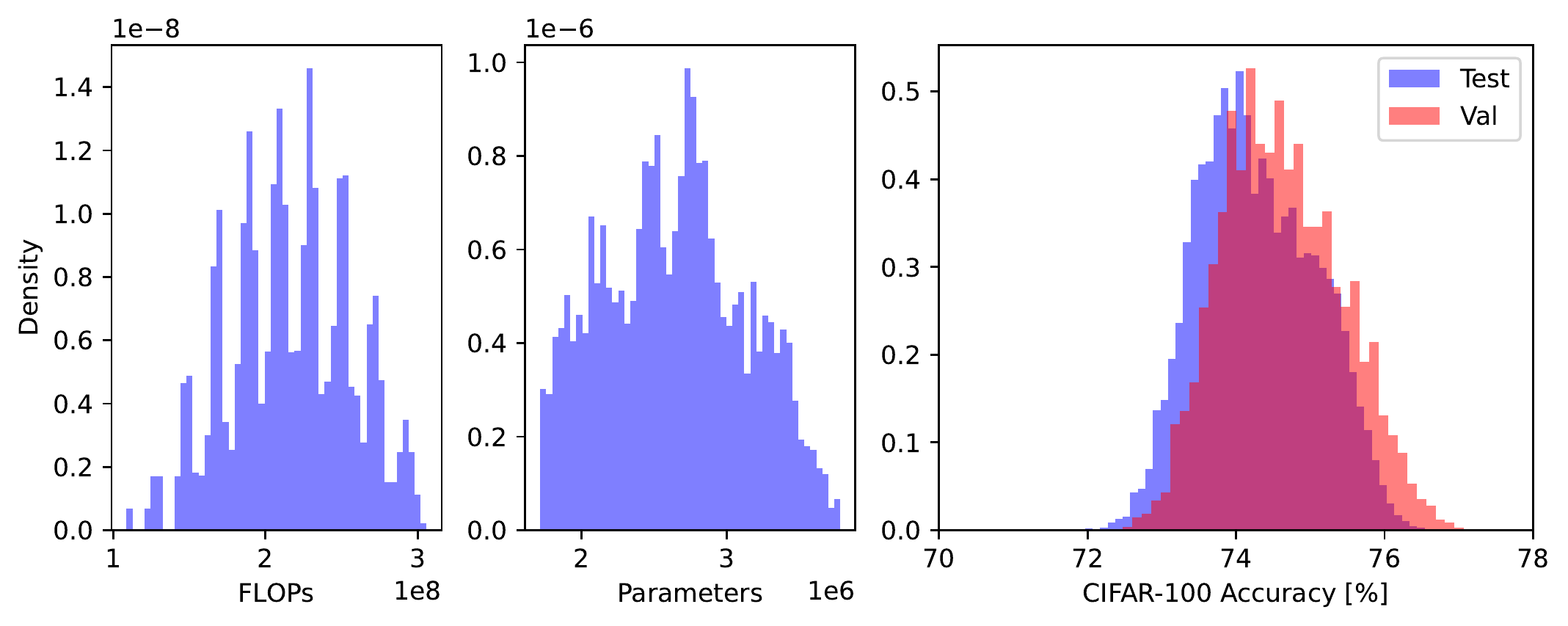}
    \caption{Distribution of FLOPs, number of parameters and accuracy in \blox{}.}
    \label{fig:basic_histograms}
    \vspace{5mm}
    \includegraphics[width=\textwidth]{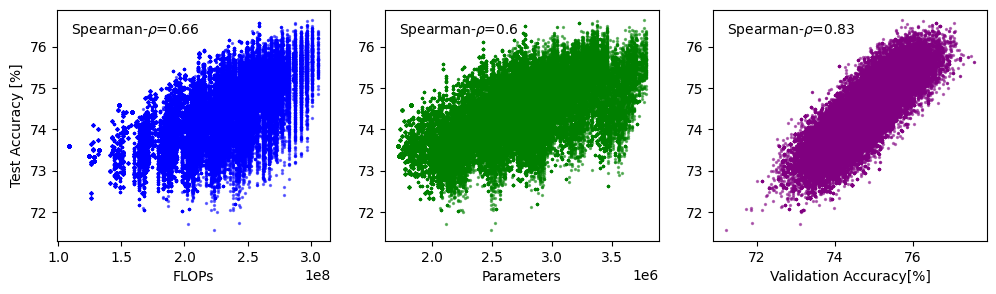}
    \caption{{(Left) Correlation between top1 test accuracy and FLOPS; (Middle) Correlation between top1 test accuracy and number of parameters; (Right) Correlation between top1 validation and test accuracy.}}
    \label{fig:correl_with_test}
\end{figure}

\noindent \textbf{FLOPs, number of parameters and accuracy.} 
Figure~\ref{fig:basic_histograms} shows the distribution of FLOPs, number of paramters, as well as top1 validation and test accuracy in \blox{}.
Regarding the relationship among these metrics,
Figure~\ref{fig:correl_with_test} (left, middle) suggests that neither FLOPs nor the number of parameters are strong factors to determine the accuracy of models.
Furthermore, the result in Figure~\ref{fig:correl_with_test} (right) shows a strong correlation between top1 validation accuracy and top1 test accuracy for all models.


\begin{figure}
    \centering
    \includegraphics[width=0.18\textwidth]{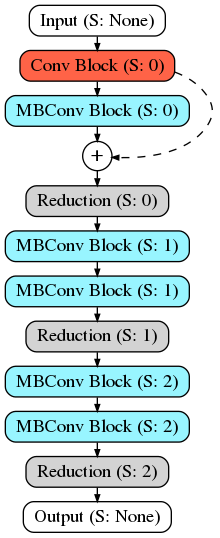}
    \includegraphics[width=0.18\textwidth]{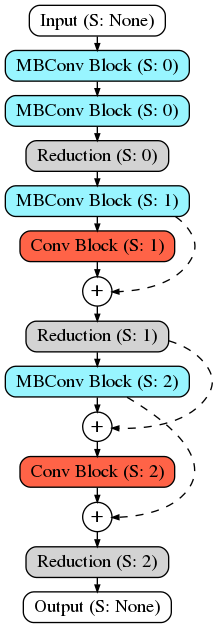}
    \includegraphics[width=0.18\textwidth]{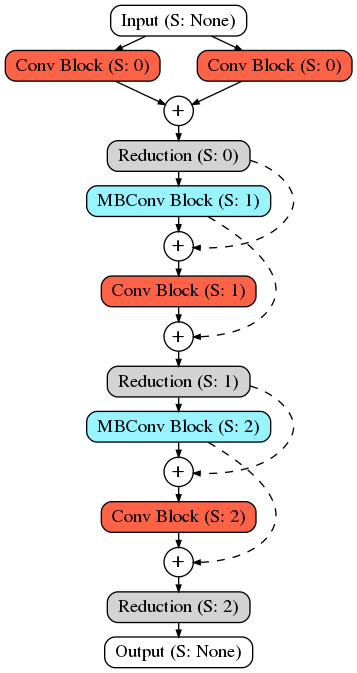}
    \includegraphics[width=0.25\textwidth]{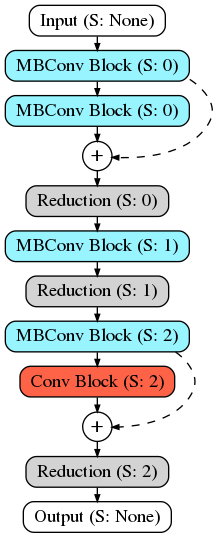}
    \includegraphics[width=0.18\textwidth]{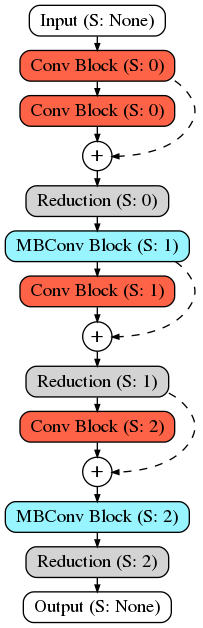}
    \caption{Best 5 models in the~\blox{} search space.}
    \label{best_models}
\end{figure}

\begin{figure}
    \centering
    \includegraphics[width=0.18\textwidth]{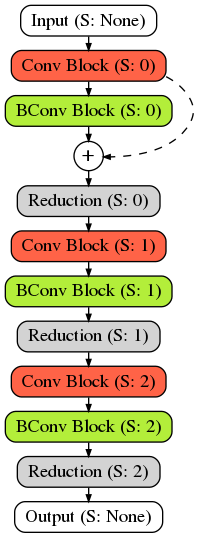}
    \includegraphics[width=0.18\textwidth]{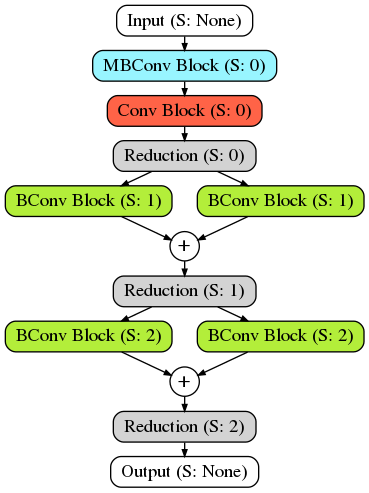}
    \includegraphics[width=0.13\textwidth]{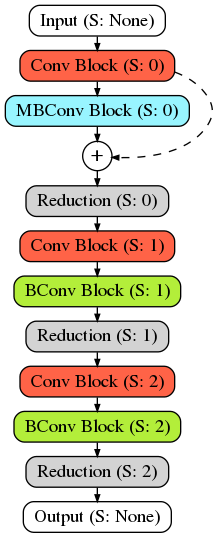}
    \includegraphics[width=0.30\textwidth]{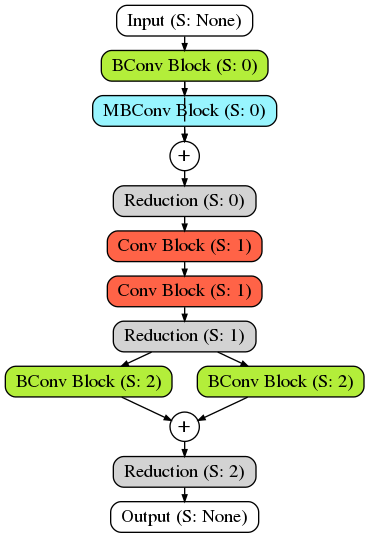}
    \includegraphics[width=0.18\textwidth]{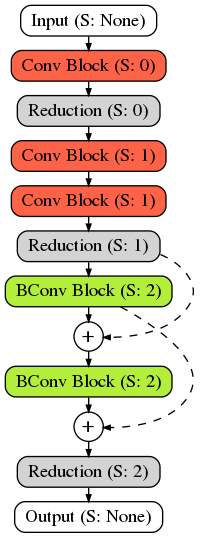}
    \caption{Worst 5 models in the~\blox{} search space.}
    \label{worst_models}
\end{figure}

\noindent \textbf{Block architectures.}
Figure~\ref{best_models} and Figure~\ref{worst_models} (at the end of this Appendix) show the models that achieve the best top1 test accuracy and the worst top1 test accuracy, respectively.
We observe that the architectures of the best models often consist of MBConv blocks whereas those of the worst models consist of BConv blocks.

\subsection{More questions about blockwise NAS}

In the main paper, we asked questions related to blockwise NAS and performed analysis using the \blox{} benchmark.
Here we ask a couple more questions to further understand the impact of predictor and distillation strategy.

\question{Does a poor predictor performance imply poor NAS results?}
\label{q:signatures_vs_nas_performance}
Correlation with accuracy is often used in NAS work as a representative proxy for the algorithm's final performance, and not without a good reason.
However, in general it is possible that a NAS algorithm might still produce good results despite being poorly correlated, as highlighted in prior work~\cite{brpnas2020}.
This is why we measure the performance of DONNA and LANA before drawing conclusions about their efficacy.
Figure~\ref{fig:ft_different_epochs} compares blockwise methods in the \blox{} search space and the models are fine-tuned by a good teacher for 10, 50, or 200 epochs.
We use \textit{FT$\alpha$} to denote fine-tuning for $\alpha$ number of epochs.
One unit of training cost is equal to the time to train for 200 epochs.
The search curves starts at $x=9$ because of the cost of block distillation (60 blocks $\times$ 3 stages $\times$ 10 epochs / 200).
Notably, the number of models trained by different approaches at a common $x$ depends on the number of fine-tuning epochs -- e.q., 90, 360 and 1,800 models are evaluated for the FT200, FT50 and FT10 settings, respectively, when the cost is 100.
Figure~\ref{fig:ft_different_epochs} shows that, with enough fine-tuning, good accuracies can be achieved with blockwise methods on the \blox{} dataset.
In the main paper, Section~\ref{sec:efficient_search} already suggested that \textbf{the type of signature used when guiding a search has a secondary role and, in general, NAS outcome is not strongly affected by it.}
Specifically, we can see that \textbf{the number of fine-tuning epochs has much more profound effect than differences in searching algorithms.}

\figvsh{0.8}{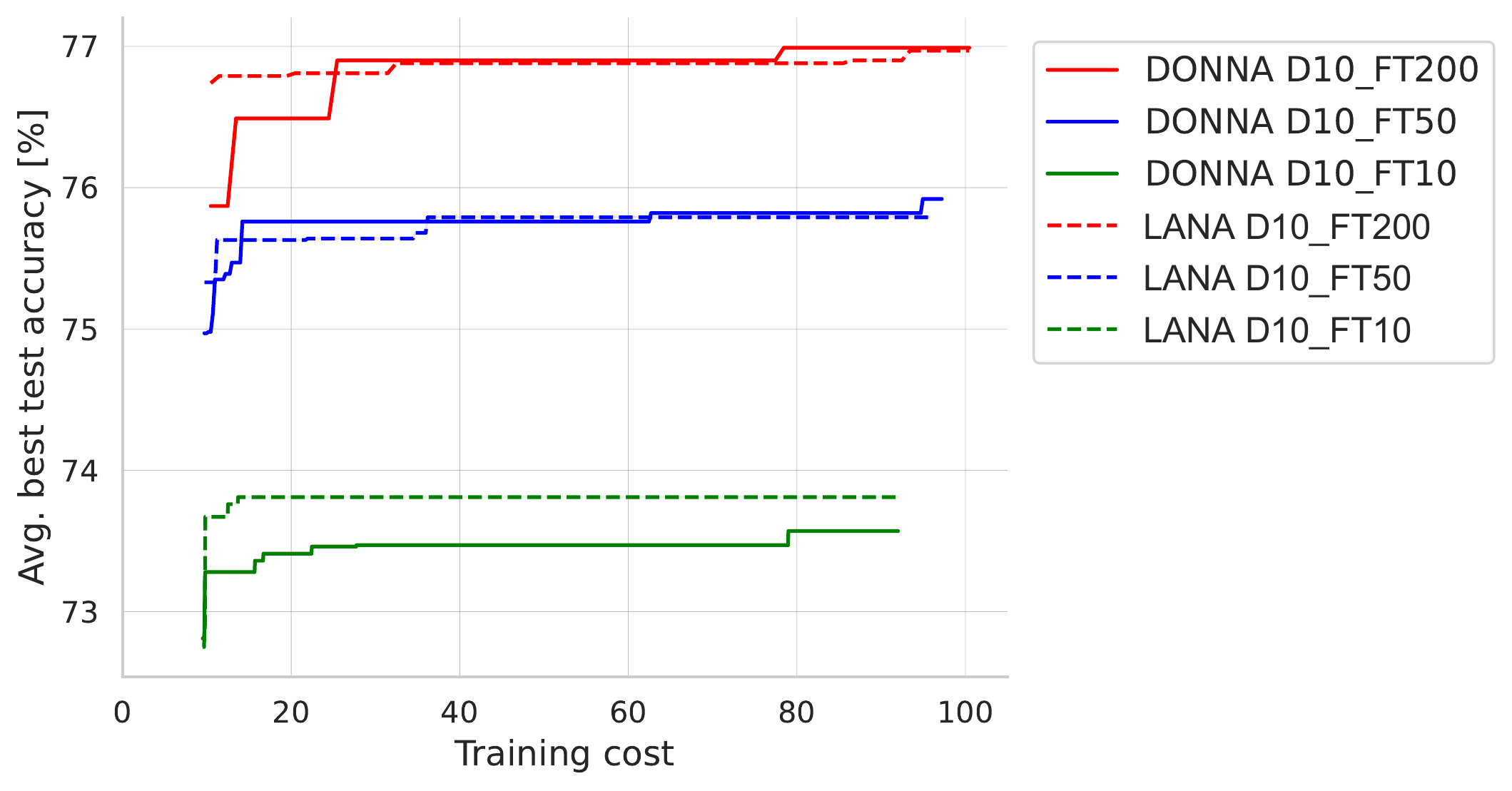}{trim = 0 20 0 20}{Comparison of NAS methods using different fine tuning epochs and M1 as the teacher. D$\alpha$ indicates distilling for $\alpha$ epochs. Similarly, FT$\alpha$ indicates fine-tuning the model for $\alpha$ epochs.}{fig:ft_different_epochs}

\question{Do we have to distill for 10 epochs?}
\label{q:dist_epochs}
Figure~\ref{fig:d_different_epochs} compares blockwise methods in which the blocks are distilled from a good teacher for 1 or 10 epochs and fine-tuned for 200 epochs.
Our results show that distillation for only 1 epoch leads to worse predictors, as indicated by the bad starting point in DONNA and slow progress in LANA.
On the other hand, distilling for 10 epochs leads to more efficient search.
This align with the results in Figure~\ref{fig:predictors} which shows that the predictor performance improves with the number of block distillation epochs.
Note that in the original DONNA paper, only 1 epoch was successfully used in blockwise distillation. 
We anticipate that this works as the original paper uses ImageNet where an epoch is significantly larger than CIFAR-100. 
This suggests that \textbf{the number of distillation epochs should be tuned based on the dataset.}

\figvs{0.8}{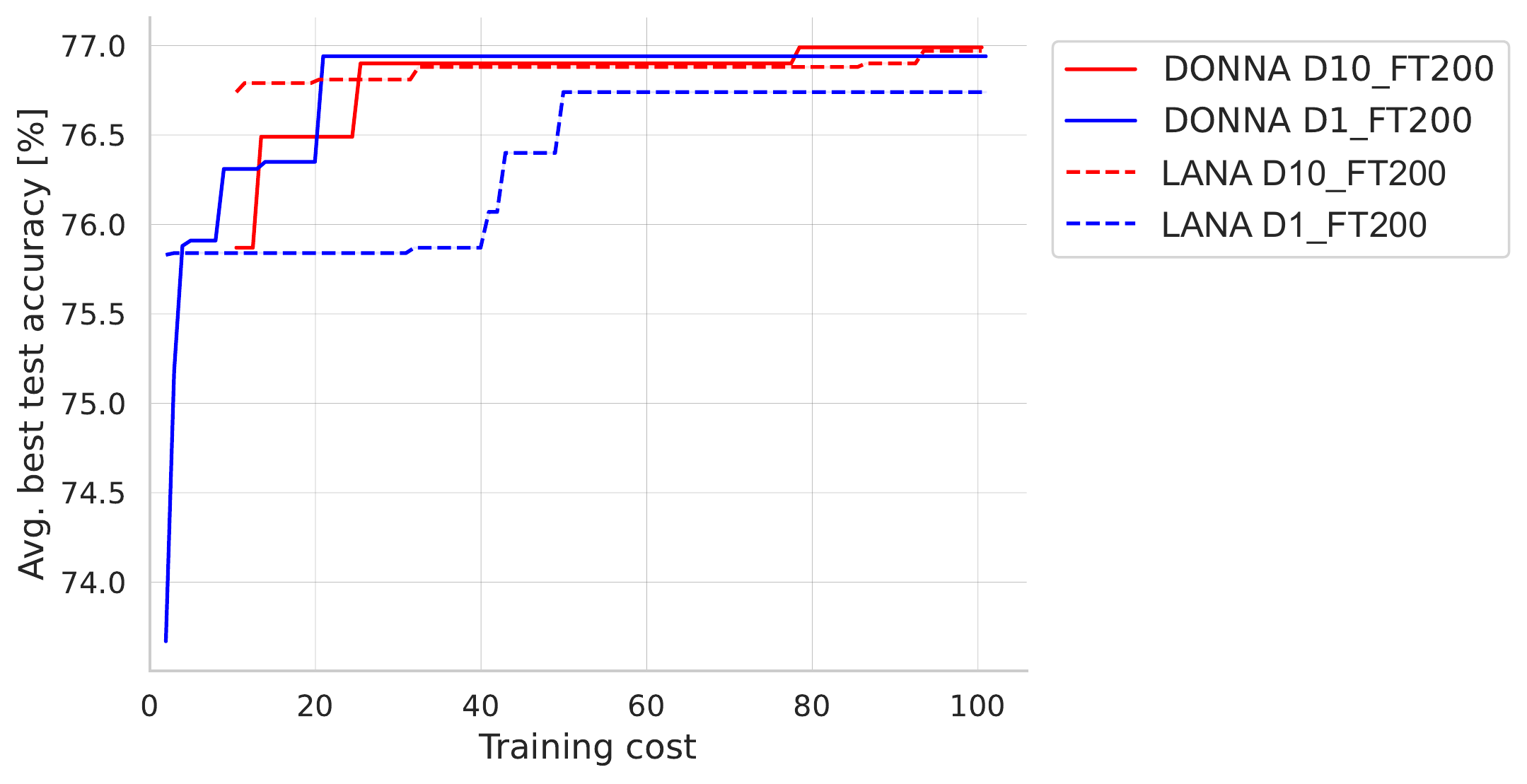}{trim = 0 0 0 0}{Comparison of NAS methods using different distillation epochs. M1 is used as the teacher.}{fig:d_different_epochs}

\subsection{Iterative fine-tuning}
As we can see from the main paper (the blue curve in Figure~\ref{fig:varying_ft_teacher}), the iterative approach has significantly improved the model accuracy without knowing a good teacher in advance.
Table~\ref{tab:iterative_ft} further shows how the selected models improve over iterations.
After the search completes (\textbf{step 1} in Question~\ref{q:find_good_teacher} of the main paper), the top-5 models are selected, as indicated by m1 to m5, and the accuracies are shown in the first row of the table.
Then the models are trained from scratch using the normal settings (\textbf{step 2}), and the results are shown in the second row.
The best model (m1 in this example) is used as the teacher to fine-tune model m1 to m5 (\textbf{step 3}), with the results in the third row.
Lastly, m4 becomes the teacher to fine-tune the models (\textbf{step 4}), the results in the forth row show that m4 achieves the best accuracy among m1 to m5.

\begin{table}[!h]
\centering
\caption {Results of iterative fine-tuning.}
\begin{tabular}{p{0.22\textwidth}p{0.10\textwidth}p{0.10\textwidth}p{0.10\textwidth}p{0.10\textwidth}p{0.10\textwidth}}
\toprule
& m1 & m2 & m3 & m4 & m5 \\
\midrule
Searched model & 70.58 & 70.50 & 70.52 & 70.61 & \textbf{70.67} \\ 
Train from scratch & \textbf{74.89} & 74.58 & 73.28 & 73.89 & 73.50 \\ 
Fine tuning iter. 1 & 76.35 & 76.27 & 75.63 & \textbf{76.36} & 75.91 \\ 
Fine tuning iter. 2 & 75.61 & 75.44 & 75.22 & \textbf{76.67} & 75.82 \\ 
\bottomrule
\end{tabular}
\label{tab:iterative_ft}
\end{table}

\subsection{Comparison of different NAS methods}

\noindent \textbf{With fine-tuning.}
Figure~\ref{fig:all_distilled} compares conventional and blockwise methods in the \blox{} search space.
We also add BRP-NAS, regularized evolution, random search and DONNA-GCN (which replaces the linear regression model by a GCN) with the distillation and fine-tuning training methodology.
We can see that all NAS methods converge to a high accuracy region when fine-tuned using a good teacher, though DONNA is still marginally the best one.
Crucially, Figure~\ref{fig:all_distilled} highlights that fine-tuning plays a bigger role in the final searched model accuracy than does the student model architecture itself.
This begs the question of whether we should focus on searching for a good teacher model architecture then use it, through distillation, to train different student models that are tailored for different target platforms.

\figvsw{0.95}{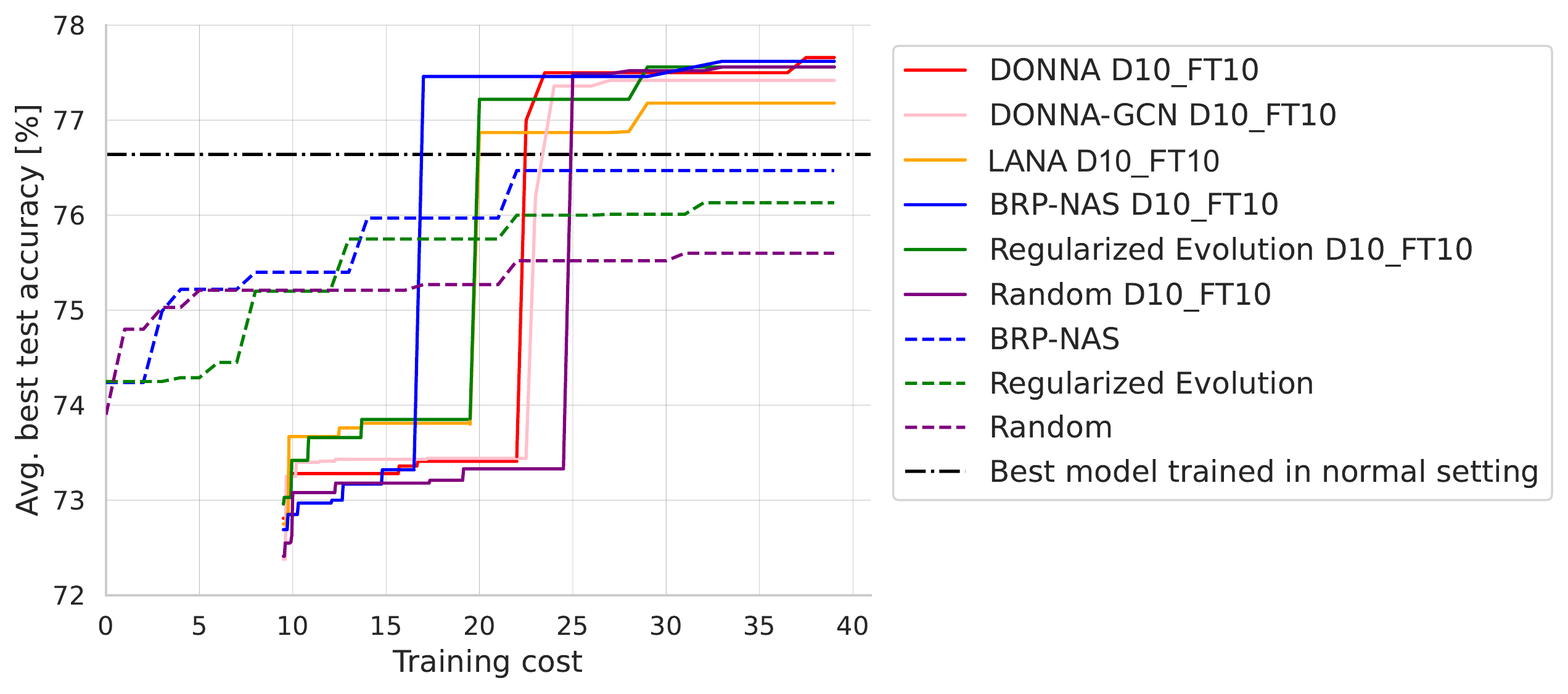}{trim = 0 0 0 0}{Comparison of conventional and blockwise methods in the \blox{} search space. For the blockwise methods, whenever the accuracy plateaus for 6 cost units, we switch to full retraining (ranking the models searched and fine-tuning them for 200 epochs using M1 as teacher) and hence the accuracy is boosted.}{fig:all_distilled}

\noindent \textbf{Without fine-tuning.}
In order to investigate the effect of fine-tuning on blockwise NAS, we run DONNA and LANA without applying any fine-tuning methodologies.
Specifically, during search, the models are trained in normal setting that blocks are not initialized with pre-trained blocks and no knowledge is distilled from a teacher.
Only block signatures (obtained by 10 epochs of distillation) are used to in the predictors to guide the search algorithms.
Figure~\ref{fig:none_distilled} shows that DONNA without fine-tuning still exhibit comparable results with conventional NAS approaches, whereas LANA struggles to show promising progress.
The results align with our previous finding that the block signature of LANA is relatively bad in predicting model performance.

\figvsh{0.55}{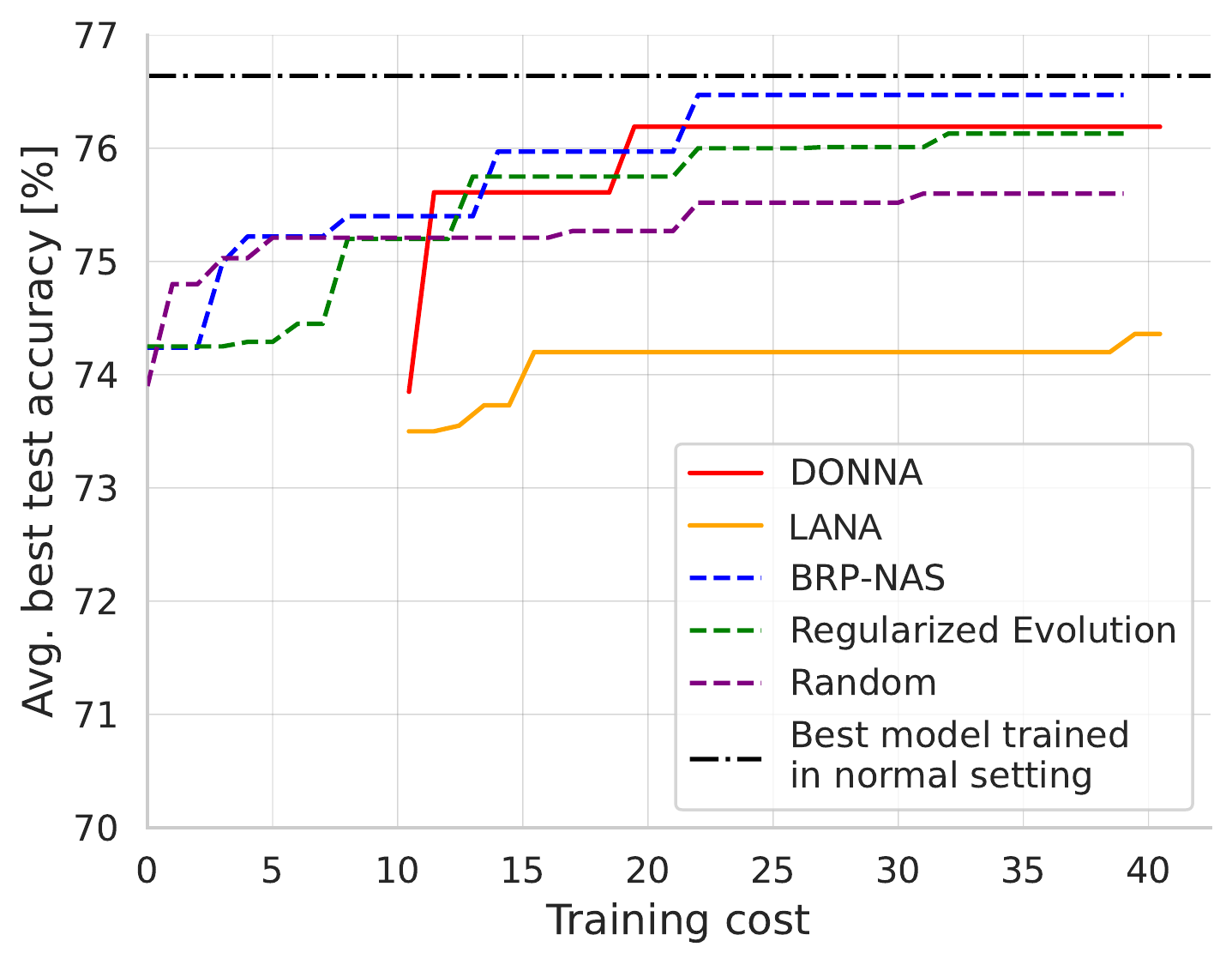}{trim = 0 0 0 0}{Comparison of blockwise methods in the \blox{} search space.}{fig:none_distilled}

\subsection{Comparison with other NAS benchmarks}
Figure~\ref{fig:blox-vs-nb2} plots Blox (macro search space, 91125 models) with NAS-Bench-201 (cell-based search space, 15625 models) as both have been trained on CIFAR-100. We also include NATS-Bench-SSS (32768 models) which is based on NAS-Bench-201 but scales the architecture size rather than topology. This figure compares Blox to the other search spaces with the same order of magnitude in terms of the number of architectures, parameters and FLOPs.

\figvsh{0.9}{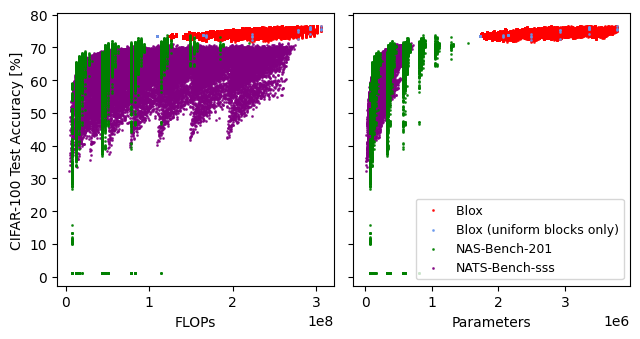}{trim = 0 0 0 0}{Comparison of \blox{} models with other NAS benchmarks.}{fig:blox-vs-nb2}

\section{Discussion and limitations} \label{sec:limitations}

\subsection{Choice of hyperparameters.}
When training our models, we decided on hyperparameters by taking 5 random models and performing a grid search over different augmentations, learning rates, optimizers, and LR schedulers.
We then picked the configuration that yielded the best results and kept using it throughout our work to provide comparable setting.
Although in our sample of 5 models we did not observe significant changes in the the relative ranking of the models when different hyperparameters were used (the only changes were happening if the models were already very close in performance), in  general it is known that different models might prefer different hyperparameters, thus making NAS results dependant on the training scheme used.
Ideally, each model would be trained with its own optimal hyperparameters.
However, this would be computationally infeasible and hence we resort to the setting that achieves the best average performance on the small sample of models, as described above.

In the case of distillation, we used the same principle by selecting a configuration giving the best average performance when the same 5 models are used in a self-distillation-like manner (i.e., the same architecture is used as a teacher and a student).
In the case of blockwise distillation, the metric that was used to score each training scheme was the average loss of distilled blocks (3 for each model).
In any other case, validation accuracy was used.
The resulting hyperparameters are summarized in Table~\ref{tab:training_hyperm}.


\subsection{Multi-objective NAS.} 
Both DONNA and LANA are originally designed to perform multi-objective NAS where accuracy should be balanced out with latency of different models.
However, we focus our analysis on the simpler case where maximizing accuracy is the sole objective.
Because of that, we have to make necessary changes in both algorithms -- for DONNA, the sub-component responsible for running NSGA-II is configured to behave like a standard single-objective evolutionary algorithm.
For LANA, budget and cost of each block ware both set to values that guarantee that the constraint in the ILP problem can never be violated.
We consider both those changes to be sensible in the context of our work.

\subsection{Potential societal impact} \label{sec:impact}
We propose a new benchmark for NAS, aiming to make NAS faster and more accessible for researchers to run generalizable and reproducible NAS experiments. 
The use of tabular NAS benchmarks allow researchers to vastly reduce the carbon footprint of traditionally compute-expensive NAS methods.
Our work can facilitates NAS research that may be have positive societal impacts (e.g., avoid training excessive amount of models) or negative societal impacts (e.g., lead to models that have fairness and security concerns).

\subsection{Discussion on previous reviews}

\noindent \textbf{Definition of macro search space.}
Previous reviewers have asked about the differences between the macro search space and cell-based search space.
We clarify the fundamental differences here --
\begin{itemize}[noitemsep,topsep=0pt,parsep=0pt,partopsep=0pt]
\item Macro search space -- each stage of a model is allowed to have different block architecture. 
In \blox{}, there are 45 unique blocks per stage, so the size of the search space is $45^3 = 91,125$.
\item Previous cell-based work such as NAS-Bench-101~\cite{nasbench1}, NAS-Bench-201~\cite{nasbench2} and DARTS~\cite{darts} -- the same cell/block is repeated to form a model. 
If we followed such cell-based setting, the size of the search space would be 45 only.
\end{itemize}

\noindent \textbf{Motivations.}
Regrading the motivation of creating this benchmark dataset, we emphasize that previous literature has mostly focused on cell-based designs.
NAS algorithm can only search for operations and connections of a cell that is repeatedly stacked within a predefined skeleton.
As shown in Figure~\ref{fig:micro_vs_macro} of the main paper, a macro search space enables layer diversity, and contains higher performing models than a cell-based search space.
Although macro search space is promising, the macro search space size grows exponentially with the number of blocks, posing a challenge to existing search algorithms.
Blockwise search algorithms are emerging (particulary DONNA and DNA that are studied in our paper), however, different methods are not comparable to each other due to different training procedures and search spaces.
As a result, we present the first large-scale benchmark on macro search space, which enables efficient ways to study NAS in this challenging setting.


\noindent \textbf{Findings.}
To address comments about the findings of the benchmark, we asked a series of questions in order to isolate relevant behaviour of the studied algorithms -- 
1) \textit{fine-tuning} (improvement brought by fine-tuning, the impact of teacher, correlation between fine-tuned models and conventionally trained models);
2) \textit{predictor} (How do block signature and end-to-end predictor affect the performance of blockwise NAS methods?);
3) \textit{search efficiency} (Can we fine-tune end-to-end model and distill blocks with few epochs? Can we fine-tune with bad teacher?).

The answers lead to a consistent conclusion -- performance is improved on blockwise methods over conventional algorithms.
In summary, 1) when a good teacher is used, blockwise NAS achieves better results (i.e. more accurate model and lower search cost) than conventional NAS ; 
2) when a bad teacher is used, we proposed a simple iterative strategy which allows us to again dominate conventional NAS.
3) efficiency of blockwise NAS is improved by utilizing reduced fine-tuning proxy followed by full fine-tuning, which is our contribution stemming from Section~\ref{sec:algo}.

\section{Dataset documentation}
Here we answer the questions outlined in the datasheets for datasets paper~\cite{datasheet_dataset}.

\subsection{Motivation}

\noindent \textbf{For what purpose was the dataset created?}
Standardized NAS benchmarks have been created to facilitate a fair comparison of NAS algorithms. 
Macro NAS, which enables the individual search for each block in a DNN, has been a promising alternative to conventional cell-based NAS.
However, macro NAS is exorbitantly expensive because the search space size grows exponentially with the number of blocks.
We present \blox{} as a large-scale benchmark to enable the empirical analysis of NAS algorithms on macro search space and to shed some light on how to perform efficient NAS in this challenging setting.

\noindent \textbf{Who created the dataset (e.g., which team, research group) and on behalf of which entity (e.g., company, institution, organization)?}
The Automated Machine Learning Group in Samsung AI Center Cambridge created the dataset.

\noindent \textbf{Who funded the creation of the dataset?}
Samsung AI Center Cambridge funded the creation of the dataset.

\subsection{Composition}

\noindent \textbf{What do the instances that comprise the dataset represent (e.g., documents, photos, people, countries)?}
Each instance in the dataset represents a network in the \blox{} search space, and the corresponding accuracies, training time and environment, number of FLOPS and parameters, and inference latencies on CIFAR-100 dataset.

\noindent \textbf{How many instances are there in total (of each type, if appropriate)?}
The dataset consists of 91,125 instances.

\noindent \textbf{Does the dataset contain all possible instances or is it a sample (not necessarily random) of instances from a larger set?}
The dataset contains all possible instances defined in the \blox{} search space.

\noindent \textbf{What data does each instance consist of?}
Each instance consists of the accuracies, training time and environment, number of FLOPS and parameters, and inference latencies on CIFAR-100 dataset.
Specifically, an instance in the base dataset contains top1 validation accuracy for all epochs and the final test top1 accuracy, as well as the training time for each model.
An instance in the extended dataset includes loss (training, validation, test), top1 training accuracy for all epochs, and top5 training accuracy for all epochs.
An instance also includes FLOPs, number of parameters, latency and architecture vector, training environment of the model.

\noindent \textbf{Is there a label or target associated with each instance?}
Each instance is associated with accuracies, training time, number of FLOPS and parameters, and inference latencies.

\noindent \textbf{Is any information missing from individual instances?}
No.

\noindent \textbf{Are relationships between individual instances made explicit (e.g., users’ movie ratings, social network links)?}
Not applicable. Each instance stands on its own.

\noindent \textbf{Are there recommended data splits (e.g., training, development/validation, testing)?}
No. It depends on the use case, for instance specific settings of the NAS alogrithms or performance predictors.

\noindent \textbf{Are there any errors, sources of noise, or redundancies in the dataset?}
No.

\noindent \textbf{Is the dataset self-contained, or does it link to or otherwise rely on external resources (e.g., websites, tweets, other datasets)?}
The dataset is self-contained and does not rely on other datasets. 
The dataset is trained on CIFAR-100, however, user can re-train the models in the \blox{} search space on different datasets.

\noindent \textbf{Does the dataset contain data that might be considered confidential (e.g., data that is protected by legal privilege or by doctor-patient confidentiality, data that includes the content of individuals’ non-public communications)?}
No.

\noindent \textbf{Does the dataset contain data that, if viewed directly, might be offensive, insulting, threatening, or might otherwise cause anxiety?}
No.





\subsection{Collection process}

\noindent \textbf{How was the data associated with each instance acquired?}
We train all instances in the \blox{} search space on CIFAR-100 dataset.
Training and validation metrics (loss and accuracy) are logged at each epoch.
Test metrics and training time are logged at the end of training for the best model. 
We also save static information consisting of FLOPs, the number of parameters, latency and architecture vector.
In addition, information about training environment are included for reproducibility. 
That includes random seed, GPU used, versions of the OS, codebase, driver, etc.

\noindent \textbf{What mechanisms or procedures were used to collect the data (e.g., hardware apparatus or sensor, manual human curation, software program, software API)?}
The data are collected by training all the models in the \blox{} search space using the source code provided in the repository.

\noindent \textbf{If the dataset is a sample from a larger set, what was the sampling strategy (e.g., deterministic, probabilistic with specific sampling probabilities)?}
Not applicable.

\noindent \textbf{Who was involved in the data collection process (e.g., students, crowdworkers, contractors) and how were they compensated (e.g., how much were crowdworkers paid)?}
The data are collected automatically by the script provided in the repository.

\noindent \textbf{Over what timeframe was the data collected?}
The initial version of data was collected between October 2021 and June 2022.

\noindent \textbf{Were any ethical review processes conducted (e.g., by an institutional review board)?}
No.


\subsection{Preprocessing}

\noindent \textbf{Was any preprocessing/cleaning/labeling of the data done (e.g., discretization or bucketing, tokenization, part-of-speech tagging, SIFT feature extraction, removal of instances, processing of missing values)?}
The weights of the trained models are not included due to limitation of data storage.

\noindent \textbf{Was the “raw” data saved in addition to the preprocessed/cleaned/labeled data (e.g., to support unanticipated future uses)?}
No.

\noindent \textbf{Is the software used to preprocess/clean/label the instances available?}
Yes, the software is available in the repository.

\subsection{Uses}

\noindent \textbf{Has the dataset been used for any tasks already?}
The dataset has been used in this paper to evaluate the performance of different NAS algorithms.

\noindent \textbf{Is there a repository that links to any or all papers or systems that use the dataset?}
Yes, it is listed in the repository.

\noindent \textbf{What (other) tasks could the dataset be used for?}
We believe that this dataset will allow researchers to evaluate the performance of different NAS algorithms, particularly blockwise methods.
In addition, the latency measurements in the dataset will enable NAS targeting different hardware devices.

\noindent \textbf{Is there anything about the composition of the dataset or the way it was collected and preprocessed/cleaned/labeled that might impact future uses?}
No.

\noindent \textbf{Are there tasks for which the dataset should not be used?}
No.


\subsection{Distribution}

\noindent \textbf{Will the dataset be distributed to third parties outside of the entity (e.g., company, institution, organization) on behalf of which the dataset was created?}
No.

\noindent \textbf{How will the dataset be distributed (e.g., tarball on website, API, GitHub)?}
The source code is available in the repository (\url{https://github.com/SamsungLabs/blox}).
The repository also includes links to allow the users to download the dataset.

\noindent \textbf{When will the dataset be distributed?}
The dataset is distributed at the same time as the published paper.

\noindent \textbf{Will the dataset be distributed under a copyright or other intellectual property (IP) license, and/or under applicable terms of use (ToU)?}
The dataset is distributed under CC BY-NC license.

\noindent \textbf{Have any third parties imposed IP-based or other restrictions on the data associated with the instances?}
No.

\noindent \textbf{Do any export controls or other regulatory restrictions apply to the dataset or to individual instances?}
No.

\subsection{Maintenance}

\noindent \textbf{Who is supporting/hosting/maintaining the dataset?}
The dataset is supported and maintained by Samsung AI Center Cambridge. 
We host the dataset on public repository (\url{https://github.com/SamsungLabs/blox}).

\noindent \textbf{How can the owner/curator/manager of the dataset be contacted (e.g., email address)?}
The contacts of the owners are included in the paper and the documentation of the public repository.

\noindent \textbf{Is there an erratum?}
Yes, it will be included in the repository.

\noindent \textbf{Will the dataset be updated (e.g., to correct labeling errors, add new instances, delete instances)?}
Yes.
Updates will be communicated via the repository, and the dataset will be versioned.

\noindent \textbf{If the dataset relates to people, are there applicable limits on the retention of the data associated with the instances (e.g., were individuals in question told that their data would be retained for a fixed period of time and then deleted)?}
Not applicable.

\noindent \textbf{Will older versions of the dataset continue to be supported/hosted/maintained?}
Yes.

\noindent \textbf{If others want to extend/augment/build on/contribute to the dataset, is there a mechanism for them to do so?}
Yes. The source code is available to allow others extend/build on/contribute to the dataset.


\subsection{Author statement}
The authors confirm all responsibility in case of violation of rights and confirm the licence associated with the dataset.





\end{document}